\documentclass[final]{cvpr}

\usepackage{times}
\usepackage{epsfig}
\usepackage{graphicx}
\usepackage{amsmath}
\usepackage{amssymb}

\usepackage{xspace}
\usepackage{multirow}
\usepackage{comment}
\usepackage{color}
\usepackage{textcomp}
\usepackage{subcaption}
\usepackage{tabulary}
\usepackage{epigraph}
\usepackage[table]{xcolor}
\definecolor{grey}{rgb}{0.9,0.9,0.9}
\usepackage{ctable}
\usepackage{algorithm}
\usepackage{algorithmic}
\usepackage{wrapfig}
\usepackage[normalem]{ulem}

\usepackage[pagebackref=true,breaklinks=true,colorlinks,bookmarks=false]{hyperref}

\def\cvprPaperID{0000} %
\def\confYear{CVPR 2021}

\DeclareMathOperator{\bool}{bool}

\begin{document}

\title{LaneRCNN: Distributed Representations for Graph-Centric Motion Forecasting}

\author{
  Wenyuan Zeng$^{1,2}$\quad Ming Liang$^{2}$ \quad Renjie Liao$^{1,2}$ \quad Raquel Urtasun$^{1,2}$\\
 $^{1}$University of Toronto\quad $^{2}$ Uber Advanced Technologies Group \\
 \small\texttt{\{wenyuan, rjliao, urtasun\}@cs.toronto.edu,
 liangming.tsinghua@gmail.com}
}

\makeatletter
\DeclareRobustCommand\onedot{\futurelet\@let@token\@onedot}
\def\@onedot{\ifx\@let@token.\else.\null\fi\xspace}

\def\eg{\emph{e.g}\onedot} \def\Eg{\emph{E.g}\onedot}
\def\ie{\emph{i.e}\onedot} \def\Ie{\emph{I.e}\onedot}
\def\cf{\emph{c.f}\onedot} \def\Cf{\emph{C.f}\onedot}
\def\aka{a.k.a\onedot} \def\Aka{A.k.a\onedot}
\def\etc{\emph{etc}\onedot} \def\vs{\emph{vs}\onedot}
\def\wrt{w.r.t\onedot} \def\dof{d.o.f\onedot}
\def\etal{\emph{et al}\onedot}
\makeatother

\newcommand{\G}{\mathcal{G}}
\newcommand{\E}{\mathcal{E}}
\newcommand{\B}{\mathcal{B}}
\newcommand{\ROI}{\textit{LaneRoI} }
\newcommand{\RCNN}{\textbf{LaneRCNN} }

\maketitle

\begin{abstract}

Forecasting the future behaviors of dynamic actors is an important task in many robotics
applications such as self-driving.  
It is extremely challenging as actors have latent intentions 
and their trajectories are governed by complex interactions between the other actors,
themselves, and the maps.
In this paper, we propose LaneRCNN, 
a graph-centric motion forecasting model.
Importantly, relying on a specially designed graph encoder, we learn a local
lane graph representation per actor (LaneRoI) to encode its past motions and the local map topology.
We further develop an interaction module which permits efficient message
passing among local graph representations within a shared global lane graph.
Moreover, we parameterize the output trajectories based on lane graphs, a more amenable prediction parameterization.
Our LaneRCNN captures the actor-to-actor and the actor-to-map relations in a distributed and map-aware manner.
We demonstrate the effectiveness of our approach on the large-scale Argoverse Motion
Forecasting Benchmark. We achieve the \textbf{1st place} on the leaderboard and
significantly outperform previous best results.

\end{abstract}

\section{Introduction}
Autonomous vehicles need to navigate in dynamic environments in a safe and comfortable manner.  
This requires predicting the future motions of other agents to understand how the 
scene will evolve. 
However, depending on each agent's intention (e.g. turning, lane-changing), the agents' future motions can involve complicated maneuvers such as yielding, nudging, and acceleration.
Even worse, those intentions are not known a priori by the ego-robot, and agents may also change their minds based on behaviors of nearby agents. 
Therefore, even with access to the ground-truth trajectory histories of the agents,
forecasting their motions is very challenging and is an unsolved problem.

\begin{figure}[t]
\begin{center}
  \includegraphics[height=3.4cm]{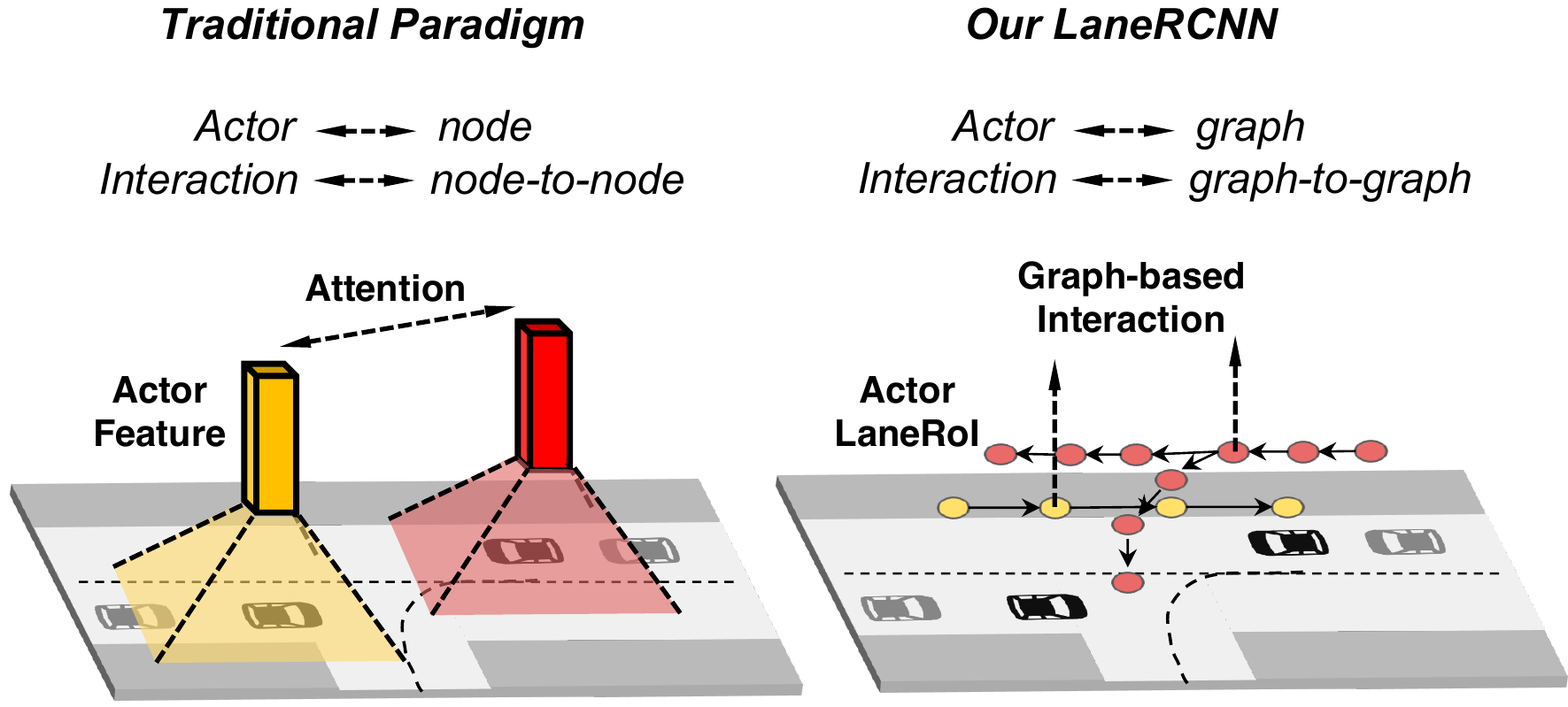}
\end{center}
\vspace{-0.2cm}
\caption{Popular motion forecasting methods encode actor and its context
information into a feature vector, and treat it as a node in an interaction graph.
In contrast, we propose a graph-based
representation \ROI per actor, which is structured and expressive. Based on
it, we model interactions and forecast motion in a map topology aware manner.}
\vspace{-0.2cm}
\label{fig:teaser}
\end{figure}

By leveraging deep learning, the motion forecasting community has been making steady progress. 
Most state-of-the-art models share a similar design principle: using a single feature
vector to characterize all the information related to an actor, as shown in Fig.~\ref{fig:teaser}, left.
They typically first encode for each actor
its past motions and the surrounding map (or other context information) into a feature
vector, which is computed either by feeding a 2D rasterization to a
convolutional neural network (CNN)
\cite{nmp,dsd,precog,chauffeurnet,covernet,intentnet}, or directly using a
recurrent neural network (RNN)
\cite{matf,mfp,vectornet,tnt,socialgan,sociallstm}.
Next, they exchange the information among actors to model interactions, \eg, via a fully-connected
graph neural network (GNN)
\cite{v2vnet,spagnn,precog,mfp,ilvm,vectornet} or an attention
mechanism \cite{interacttransformer,sophie,socialatt,carnet,mercat2020multi}. 
Finally, they predict future motions per actor from its feature vector
via a regression header \cite{lgn,mfp,precog,nmp,intentnet,pnpnet,attnmp}.

\begin{figure*}[t]
\begin{center}
  \includegraphics[height=6.0cm]{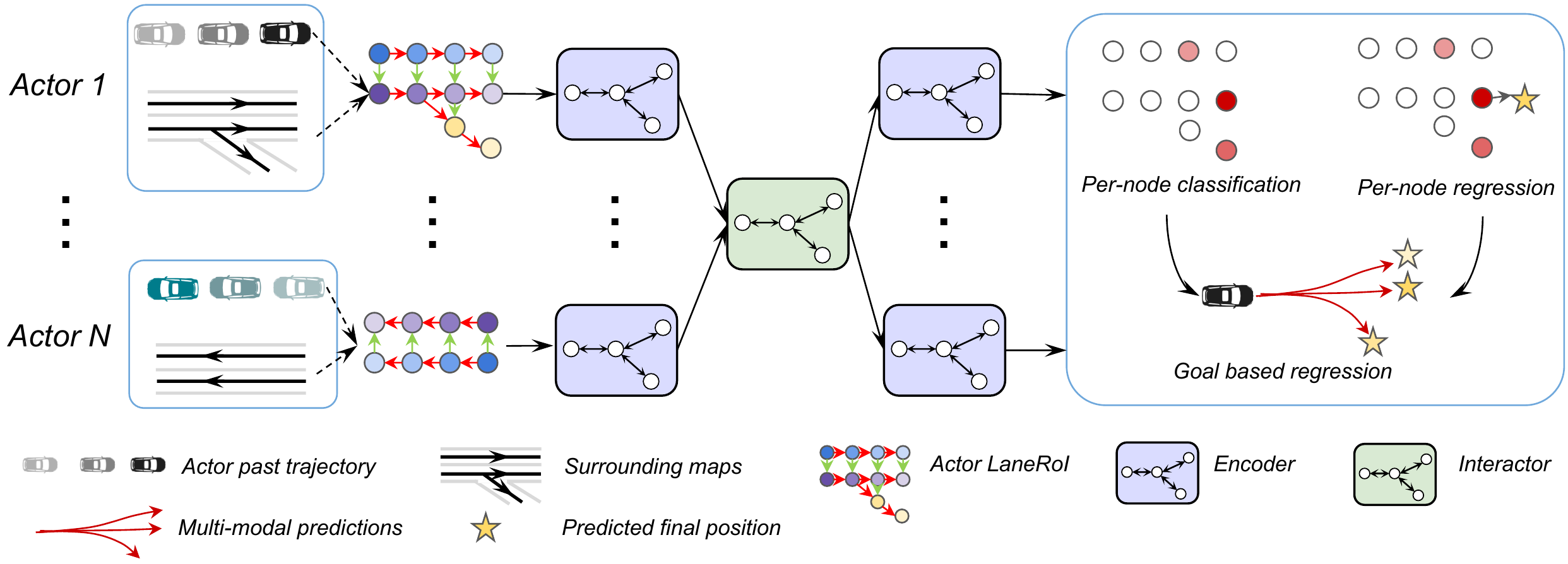}
\end{center}
\vspace{-0.3cm}
\caption{Overview of LaneRCNN. It first
encodes each actor with our proposed \ROI representation, processes
each \ROI with an encoder, and then models interactions among actors with a
graph-based interactor. Finally, LaneRCNN predicts final positions of actors in
a fully-convolutional manner, and then decodes full trajectories based on these
positions.}
\vspace{-0.1cm}
\label{fig:lanercnn}
\end{figure*}

Although such a paradigm has shown competitive results, it has three main shortcomings:
1) Representing the context information of large regions of space, such as fast moving actors traversing possibly a hundred meters within five seconds, with a single vector is difficult.
2) Building a fully-connected interaction graph among actors ignores important map
structures. For example, an unprotected left turn vehicle should yield to oncoming traffic, while two
spatially nearby vehicles driving on opposite lanes barely interact with each other.
3) The regression header does not explicitly leverage the lane information, which could provide a good inductive bias for accurate predictions. 
As a consequence, regression-based
predictors sometimes forecast \textit{shooting-out-of-road} trajectories, which
are unrealistic.

In this paper, we propose a graph-centric motion forecasting model, \ie, LaneRCNN, 
to address the aforementioned issues. 
We represent an actor and its context 
in a distributed and map-aware manner by 
constructing an actor-specific graph, called Lane-graph Region-of-Interest (\textit{LaneRoI}), along with node 
embeddings that encode the past motion and map semantics. 
In particular, we construct  \textit{LaneRoI} following the topology 
of lanes that are relevant to this actor, where nodes on this graph correspond
to small spatial regions along these lanes, and edges represent the topological and spatial relations among regions. 
Compared to using a single vector to encode all the information of a large region, our \textit{LaneRoI} naturally
preserves the map structure and captures the more fine-grained information, 
as each node embedding only needs to represent the local context within a small
region. 
To model interactions, we embed the \textit{LaneRoI}s of all actors to
a global lane graph  and then propagate the information over this global graph.
Since the \textit{LaneRoI}'s of interacting actors are highly relevant, those actors will
share overlapping regions on the global graph, thus having more frequent
communications during the information propagation compared to 
irrelevant actors. 
Importantly, this process neither requires any heuristics nor makes any oversimplified
assumptions while learning interactions conditioned on maps. 
We then predict future motions on each
\textit{LaneRoI} in a \emph{fully-convolutional} manner, such that small regions along
lanes (nodes in \textit{LaneRoI}) can serve as anchors and provide good priors
. We demonstrate the effectiveness of our method on the
large-scale Argoverse motion forecasting benchmark \cite{argoverse}. We achieve
the \textbf{first rank} on the challenging Argoverse competition leaderboard
\cite{argoleaderboard}, significantly
outperforming previous results.

\section{Related Work}

\paragraph{Motion Forecasting:}
Traditional methods use hand-crafted features and rules based on human knowledge 
to model interactions and constraints in motion forecasting
\cite{choi2013understanding,choi2012unified,deo2018would,helbing1995social,mehran2009abnormal,yamaguchi2011you,weichiuplay}, which are sometimes oversimplified and not scalable. 
Recently, learning-based approaches employ the deep learning and
significantly outperform traditional ones.
Given the actors and the scene, a deep
forecasting model first needs to design a format to encode the information.
To do so, previous methods \cite{precog,chauffeurnet,covernet} often rasterize
the trajectories of
actors into a Birds-Eye-View (BEV) image, with different channels representing
different observation timesteps, and then apply a CNN and RoI pooling
\cite{fasterrcnn,maskrcnn} to extract actor features. 
Maps can be encoded similarly
\cite{nmp,dsd,intentnet,chauffeurnet,mfp}. 
However, the square receptive fields of a CNN may not be efficient to
encode actor movements \cite{lgn}, which are typically long curves.
Moreover, the map rasterization may lose useful information like lane topologies. 
RNNs are an alternative way to encode actor kinematic information
\cite{matf,mfp,vectornet,tnt,socialgan,sociallstm} compactly and efficiently.
Recently, VectorNet \cite{vectornet} and LaneGCN \cite{lgn} generalized such compact encodings to map representations. 
VectorNet treats a map
as a collection of polylines and uses a RNN to encode them, while LaneGCN builds a
graph of lanes and conducts convolutions over the graph. Different from all these
work, we encode both actors and maps in an unified graph representation, which
is more structured and powerful.

Modeling interactions among actors is also critical for
a multi-agent system. Pioneering learning-based work design a social-pooling mechanism
\cite{sociallstm,socialgan} to aggregate the information from nearby actors.
However, such a 
pooling operation may potentially lose
actor-specific information. To address this,
attention-mechanism \cite{sophie,socialatt,carnet,sun2019relational} or GNN-based
methods \cite{dsd,interacttransformer,lgn,spagnn,precog,mfp,ilvm,vectornet} build actor
interaction graphs (usually fully-connected with all actors or k-nearest
neighbors based),
and perform attention or message passing to update actor features.
Social convolutional pooling \cite{matf,socialconvpool,pip} has also been explored, which maintains the spatial
distribution of actors. 
However, most of these work do not explicitly consider map structures,
which largely affects interactions among actors in reality. 

To generate each actor's predicted futures, many works sample multi-modal futures under a conditional variational auto-encoder (CVAE) framework
\cite{desire,r2p2,mfp,precog,ilvm}, or with a multi-head/mode regressor
\cite{lgn,cui2019multimodal,mercat2020multi}.
Others output discrete sets of
trajectory samples \cite{dsd,covernet,multipath} or occupancy maps
\cite{jain2019discrete,p3}. Recently, TNT \cite{tnt}
concurrently and independently designs a similar output parameterization as ours
where lanes are used as priors for the forecasting. 
Note that, in addition to the parameterization, we contribute a novel graph representation and a powerful
architecture which significantly outperforms their results.

\paragraph{Graph Neural Networks:}

Relying on operators like graph convolution and message passing, graph neural networks (GNNs) and their variants \cite{scarselli2008graph,bruna2013spectral,li2015gated,kipf2016semi,hamilton2017inductive,liao2019lanczosnet} generalize deep learning on regular graphs like grids to ones with irregular topologies.
They have achieved great successes in learning useful graph representations for
various tasks \cite{monti2017geometric,qi20173d,teney2017graph,li2017situation,garcia2018few}.
We draw inspiration from the general concept ``ego-graph'' and propose \textit{LaneRoI}, which is specially designed for lane graphs and captures both the local map topologies and the past motion information of an individual actor.
Moreover, to capture interactions among actors, we further propose an
interaction module which effectively communicates information among
\textit{LaneRoI} graphs.

\section{LaneRCNN}

Our goal is to  predict the future motions of all actors in a scene, given their
past motions and an HD map. Different from existing work, we represent an actor
and its context with a \textit{LaneRoI}, an actor-specific graph which is more
structured and expressive than the single feature vector used in the literature. 
Based
on this representation, we design LaneRCNN, a graph-centric motion forecasting
model that encodes context, models interactions between actors, and predicts future motions all in
a map topology aware manner. An overview of our model is shown in
Fig.~\ref{fig:lanercnn}.

In the following, we first introduce our problem formulation and notations 
in Sec. \ref{sec:notation}. We then define our \ROI representations in Sec.
\ref{sec:laneroi}. In Sec. \ref{sec:backbone}, we explain how LaneRCNN processes features and models interactions via graph-based message-passing. 
Finally, we show our map-aware trajectory prediction and learning objectives
in Sec. \ref{sec:output} and Sec. \ref{sec:learning} respectively.

\begin{figure}[t]
\begin{center}
  \includegraphics[height=3.4cm]{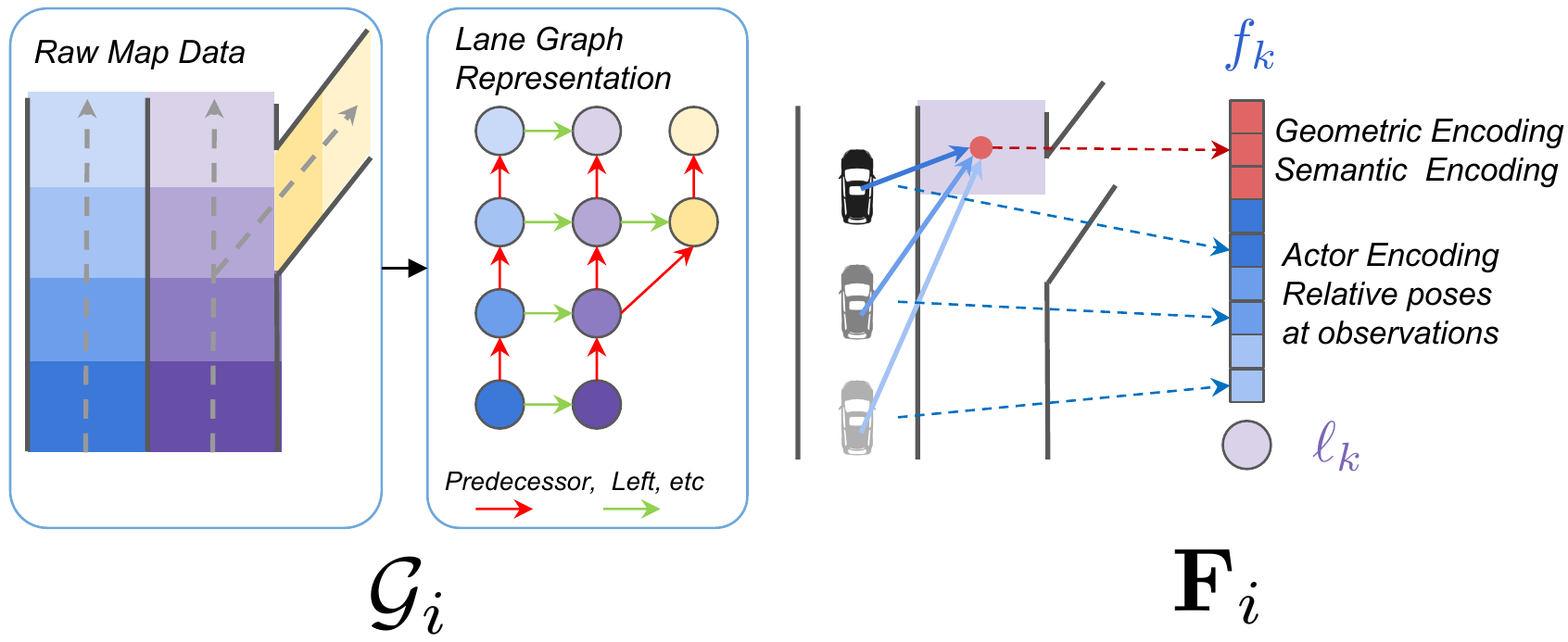}
\end{center}
\vspace{-0.2cm}
\caption{
The LaneRoI of the actor $i$ is a collection of a graph $\G_i$ (constructed following lane topology: nodes as
lane segments and edges as segment connectivities) and node
embeddings $\mathbf{F}_i$ (encoding motions of the actor, as well as
geometric and semantic properties of lane segments).}
\vspace{-0.2cm}
\label{fig:laneroi}
\end{figure}
\subsection{Problem Formulation}
\label{sec:notation}
We denote the  past motion of the $i$-th actor as a set of 2D points encoding
the center locations over the past $L$ 
 timesteps, \ie, $\left\{(x_i^{-L}, y_i^{-L}), \cdots, (x_i^{-1},
y_i^{-1})\right\}$, with $(x,y)$  the  2D coordinates  in bird's eye view (BEV). Our goal is to forecast the future motions of all actors in the scene 
$\left\{(x_i^{0}, y_i^{0}), \cdots, (x_i^T, y_i^T) | i = 1, \cdots, N\right\}$,
where $T$ is our prediction horizon and $N$ is the number of actors. 

In addition to the past kinematic information of the actors, maps also play an important role 
for motion forecasting since (i) actors usually follow lanes on the map, 
(ii) the map structure determines the right of way, which in turns affects the interactions among actors.
As is common practice in self-driving, we assume an HD map is accessible, 
which contains lanes and associated semantic attributes, \eg,
turning lane and lane controlled by traffic light.  
Each lane is composed of
many consecutive lane segments $\ell_i$, which are short segments
along the centerline of the lane.
In addition, a lane segment $\ell_i$ can have pairwise relationships with
another segment $\ell_j$ in the same lane or in another lane, 
such as $\ell_i$ being a successor of $\ell_j$ or a left neighbor.

\subsection{LaneRoI Representation}
\label{sec:laneroi}

\paragraph{Graph Representation:}
One straight-forward way to represent an actor and its context (map) information
is by first rasterizing both its trajectory as well as the map to form a 2D BEV image, and then cropping the underlying representation centered in the actor's location in BEV \cite{dsd, mfp, matf, spagnn}.
However, rasterizations are prone to information loss such as 
connectivities among lanes. 
Furthermore, it is a rather inefficient representation
since actor motions are expanded typically in the direction along the lanes, not across them. 
Inspired by \cite{lgn}, we instead use a graph representation for our \ROI to
preserve the structure while being compact. For each actor $i$ in the scene, we first
retrieve all relevant lanes that this actor can possibly go to in the prediction horizon
$T$ as well as come from in the observed history horizon $L$. We then convert the
lanes into a directed graph $\G_i =
\{\mathcal{V}, \{ \mathcal{E}_\text{suc}, \mathcal{E}_\text{pre},
    \mathcal{E}_\text{left}, \mathcal{E}_\text{right} \}\}$
where each node $v \in \mathcal{V}$ represents a lane segment within those lanes 
and the lane topology is represented by different types of edges $\mathcal{E}_r$,
encoding the following relationships: predecessor, successor, left
and right neighbor.  
Two nodes are connected by an edge $e \in \mathcal{E}_r$ if the corresponding lane segments $\ell_i,
\ell_j$ have a relation $r$, \eg, lane segment $\ell_i$ is a
successor of lane segment $\ell_j$.
Hereafter, we will use the term node interchangeably with the term lane segment.

\paragraph{Graph Input Encoding:}
The graph $\G_i$ only characterizes map structures around the $i$-th actor without much information about the actor.
We therefore augment the graph with a set of node embeddings to
construct our \ROI.
Recall that each node $k$ in $\G_i$ is associated with a lane
segment $\ell_k$. We design its  embedding $f_k \in \mathbb{R}^C$ 
to capture the geometric and semantic information of
$\ell_k$, as well as its relations with the actor.
In particular, geometric features include the center location, the orientation and the
curvature of $\ell_k$; semantic features include binary features indicating if
$\ell_k$ is a turning lane,  if it is currently controlled by a red light,
\etc. To encode the actor information into $f_k$, we note that the past motion of an
actor can be identified as a set of 2D displacements, defining the 
movements between consecutive timesteps. Therefore, we also include the relative
positions and orientations of these 2D displacements \wrt $\ell_k$
into $f_k$ which encodes actor motions in a map-dependent manner.
This is beneficial for understanding actor behaviors \wrt the map, \eg, a trajectory that 
steadily deviates from one lane and approaches the neighboring lane is highly likely 
a lane change.
In practice, it is important to clamp the actor information, \ie, if $\ell_k$ is more than 5 meters away from the
actor we replace the actor motion embedding in $f_k$ with zeros. 
We hypothesize that such a restriction encourages the model to learn better representations via the message passing over the graph.
To summarize, $(\G_i, \mathbf{F}_i)$ is the \ROI of the actor $i$, encoding
the actor-specific information for motion forecasting,
where $\mathbf{F}_i \in
\mathbb{R}^{M_i \times C}$ is the collection of node embeddings $f_k$ and 
$M_i$ is the number of nodes in $\G_i$.

\subsection{LaneRCNN Backbone}
\label{sec:backbone}
\begin{figure}[t]
\begin{center}
  \includegraphics[height=4.5cm]{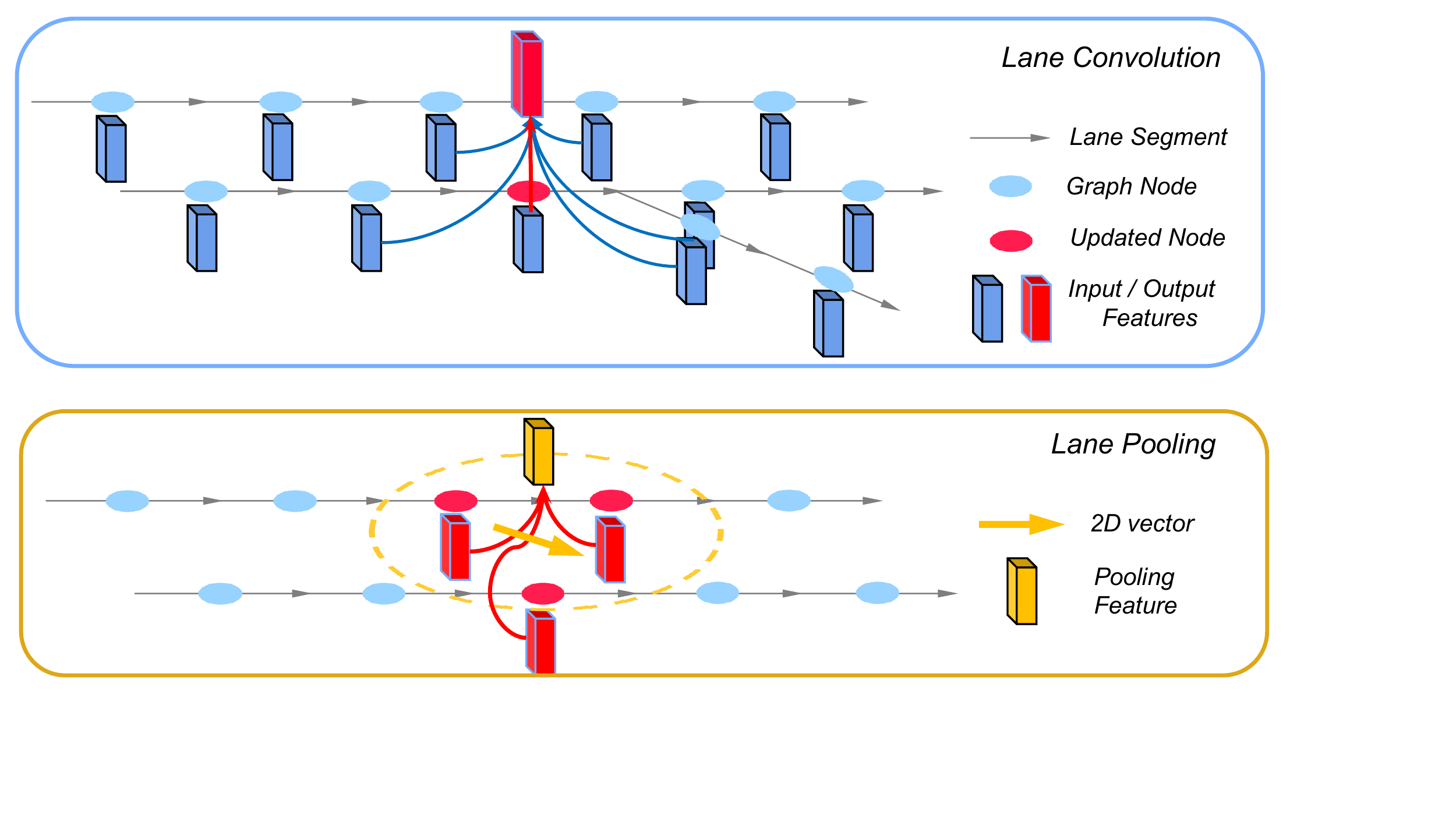}
\end{center}
\vspace{-0.2cm}
\caption{An illustration for lane convolution and lane
pooling operators, which have similar functionalities as their 2D counterparts
while respecting the lane topology.}
\label{fig:operator}
\end{figure}

As \textit{LaneRoI}s have irregular graph structures, we can not apply standard
2D convolutions to obtain feature representations. In the following, we first introduce
the lane convolution and pooling operators (Fig.~\ref{fig:operator}), which serve similar
purposes as their 2D counterparts while respecting the graph topology. 
Based on these operators, we then describe how our LaneRCNN updates features of
each \ROI as well as handles interactions among all \textit{LaneRoI}s (actors).

\paragraph{Lane Convolution Operator:}
We briefly introduce the lane convolution which was originally proposed in \cite{lgn}
Given a \ROI $(\G_i, \mathbf{F}_i)$, 
a lane convolution updates features $\mathbf{F}_i$ by aggregating features from
its neighborhood (in the graph). 
Formally, we use $\E_i(r)$ to denote the 
binary adjacency matrix for $\G_i$ under the relation $r$, \ie, the $(p, q)$ entry 
in this matrix is $1$ if lane segments $\ell_p$ and $\ell_q$ have the relation $r$ 
and $0$ otherwise. 
We denote the $n$-hop connectivity under the relation $r$ as the matrix 
$\bool\left(\E_i(r) \cdot \E_i(r) \cdots \E_i(r)\right) = \bool
\left(\E_i^n(r)\right)$, where the operator $\bool$ sets any non-zero entries to one 
and otherwise keeps them as zero. 
The output node features are updated as follows,
\begin{equation}
  \label{eq:conv}
  \mathbf{F}_i \leftarrow \Psi \left( \mathbf{F}_i\mathbf{W} + \sum_{r, n}
  \bool \left(\E_i^n(r)\right)\mathbf{F}_i\mathbf{W}_{n, r} \right),
\end{equation}
where both $\mathbf{W}$ and $\mathbf{W}_{n, r}$ are learnable
parameters, $\Psi(\cdot)$ is a non-linearity consisted of
LayerNorm \cite{layernorm} and ReLU \cite{relu},
and the summation is over all possible relations $r$
and hops $n$. In practice, we use 
$n \in \left\{1, 2, 4,
8, 16, 32\right\}$.
Such a multi-hop mechanism mimics the dilated convolution \cite{yu2015multi} and effectively enlarges
the receptive field.

\paragraph{Lane Pooling Operator:}
We design a lane pooling operator which is a learnable pooling function. Given a \ROI
$(\G_i, \mathbf{F}_i)$, recall $\G_i$ actually corresponds to a number of
lanes spanned in the 2D plane (scene). For an arbitrary 2D vector
$\mathbf{v}$ in the plane, a lane pooling operator pools, or `interpolates',
the feature of $\mathbf{v}$ from $\mathbf{F}_i$. Note that
$\mathbf{v}$ can be a lane segment in another graph $\G_j$ (spatially
close to $\G_i$). Therefore, lane pooling helps
communicate information back and forth between graphs, which we will explain in the interaction part.
To generate the feature $f_\mathbf{v}$ of vector $\mathbf{v}$, we first retrieve
its `neighboring nodes' in $\G_i$, by checking if the center distance between a lane segment
$\ell_k$ in $\G_i$ and vector $\mathbf{v}$ is smaller than a certain threshold. A naive
pooling strategy is to simply take a mean of those $\ell_k$. However, this
ignores the fact that relations between $\ell_k$ and $\mathbf{v}$ can vary a lot
depending on their relative pose: a lane segment that is perpendicular to
$\mathbf{v}$ (conflicting) and the one that is aligned with $\mathbf{v}$
have very different semantics. Inspired by the generalized convolution on graphs/manifolds
\cite{monti2017geometric, contconv, lgn}, we use the relative pose and some non-linearities to
learn a pooling function. In particular, we denote the set of surrounding
nodes on $\G_i$ as $\mathcal{N}$, and the relative pose between $\mathbf{v}$ and
$\ell_k$ as $\Delta_{\mathbf{v}k}$ which includes relative position and
orientation. The pooled feature $f_{\mathbf{v}}$ can then be written as,
\begin{equation}
  \label{eq:pool}
  f_{\mathbf{v}} = \mathcal{M}_b\left(\sum_{k\in \mathcal{N}}
    \mathcal{M}_a\left(\left[
        f_k, \Delta_{\mathbf{v}k}
\right]\right)\right),
\end{equation}
where $[\cdots]$ means concatenation and $\mathcal{M}$
is a two-layer multi-layer perceptron (MLP).

\paragraph{LaneRoI Encoder:}
Equipped with operators introduced above, we now describe how LaneRCNN processes
features for each \textit{LaneRoI}. Given a scene, we first
construct a \ROI per actor and encode its input information into node
embeddings as described in Sec. \ref{sec:laneroi}. Then, for each \textit{LaneRoI}, we apply
four lane convolution layers and get the updated node embeddings
$\mathbf{F}_i$.
Essentially, a lane convolution layer propagates information from a node to
its (multi-hop) connected nodes. Stacking more layers builds larger receptive
fields and has a larger model capacity. However, we find deeper
networks do not necessarily lead to better performances in practice, possibly due to the well-known
difficulty of learning long-term dependencies. To address this, we introduce a
graph shortcut mechanism on \textit{LaneRoI}. 
The graph shortcut layer can be applied after any layer of lane convolution:
we aggregate $\mathbf{F}_i$ output from the previous layer into a global
embedding with the same dimension as node embeddings, and then add it to embeddings
of all nodes in $\G_i$.
Recall that the actor past motions are a number of 2D vectors, \ie, movements
between consecutive timesteps. We use the lane pooling to extract
features for these 2D vectors. A 1D CNN with downsampling is then applied to these features to
build the final shortcut embedding. Intuitively, a lane convolution may suffer
from the diminishing information flow during the message-passing, while such
a shortcut can provide an auxiliary path to communicate among far-away nodes
efficiently. We will show that the shortcut significantly boosts the performance in the ablation study.

\paragraph{LaneRoI Interactor:}
So far, our \ROI encoder provides good features for a given actor, but it
lacks the ability to model interactions among different actors, which is
extremely important for the motion forecasting in a multi-agent system. We now
describe  how we handle actor interactions under \ROI representations.
After processing all \textit{LaneRoI}s with the \ROI encoder (shared weights), we build a global lane graph
$\G$ containing all lanes in the scene. Its node embeddings are constructed by
projecting all \textit{LaneRoI}s to $\G$ itself. We then apply four lane convolution layers on $\G$ to perform message passing. Finally, we distribute the
`global node' embeddings back to each \ROI. Our design is motivated by the fact
that \textit{actors have interactions since they share the same space-time region}. Similarly,
in our model, all \textit{LaneRoI}s share the same global graph $\G$ where
they communicate with each other following map structures. 

In particular, suppose we have a set of \textit{LaneRoI}s $\{(\G_i,
\mathbf{F}_i)|i=1,\cdots,N\}$ encoded from previous layers and a global lane
graph $\G$. For each node in $\G$, we use a lane pooling to construct its
embedding: retrieving its neighbors from all \textit{LaneRoI}s as $\mathcal{N}$,
measured by center distance, and then applying Eq. \ref{eq:pool}. This ensures
each global node has the information of all those actors that could 
interact with it. The distribute step is an inverse process: for each node in
$\G_i$, find its neighbors, apply a lane pooling, and add the resulted embedding to original $\mathbf{F}_i$ (serving as a
skip-connection).

\begin{table*}[t]
\vspace{-0.2cm}
\centering
\begin{tabular}{l|l|ccc|cc>{\columncolor{grey}}c}
  \specialrule{.2em}{.1em}{.1em}
 & \multirow{2}{*}{Method} & \multicolumn{3}{c|}{K=1} & \multicolumn{3}{c}{K=6} \\
 & & minADE & minFDE & MR & minADE & minFDE & MR \\
  \hline
  \multirow{3}{*}{Argoverse Baseline \cite{argoverse}} & NN & 3.45 & 7.88 & 87.0 & 1.71 & 3.28& 53.7 \\
                       & NN+map & 3.65 & 8.12 & 94.0 & 2.08 & 4.02 & 58.0 \\
                       & LSTM+map & 2.92 & 6.45 & 75.0 & 2.08 & 4.19 & 67.0 \\
  \hline
  \multirow{4}{*}{Leaderboard \cite{argoleaderboard}} & TNT (4th) \cite{tnt} & 1.78 & 3.91 & 59.7 & 0.94 & 1.54 & 13.3   \\
                               & Jean (3rd) \cite{mercat2020multi} & 1.74 & 4.24 & 68.6 & 1.00 & \textbf{1.42} & 13.1  \\
                               & Poly (2nd) \cite{argoleaderboard} & 1.71 & 3.85 & 59.6 & \textbf{0.89} & 1.50 & 13.1 \\
                               \cline{2-8}
                   & Ours-LaneRCNN (1st) & \textbf{1.69} & \textbf{3.69} &
  \textbf{56.9} & 0.90 & 1.45 &
  \textbf{12.3} \\
  \specialrule{.1em}{.05em}{.05em}

\end{tabular}
\caption{Argoverse Motion Forecasting Leaderboard. All metrics are lower the
  better and \colorbox{grey}{Miss-Rate (MR, K=6)} is the official ranking metric.}
\label{table:argo}
\vspace{-0.2cm}
\end{table*}

\subsection{Map-Relative Outputs Decoding}
\label{sec:output}

The future is innately multi-modal and an actor can take many different yet possible
future motions. Fortunately, different modalities can be largely characterized by
different goals of an actor. Here, a goal means a
final position of an actor at the end of prediction horizon. Note that actors
mostly follow lane structures and thus their goals are usually close to a lane
segment $\ell$. Therefore, our model can predict the final goals of an actor in a
fully convolutional manner, based on its \ROI features. Namely, we apply a
2-layer MLP on each node feature $f_k$, and output five values including the
probability that $\ell_k$ is the closest lane segment to final destination 
$p(\ell_k=\text{goal})$, as well as relative residues from $\ell_k$ to the final
destination $x_{gt} - x_{k}$, $y_{gt} - y_{k}$, $\sin(\theta_{gt} - \theta_k)$,
$\cos(\theta_{gt} - \theta_k)$.

Based on results of previous steps, we select the top K\footnote{On
Argoverse, we follow the official metric and use K=6. We also remove duplicate
goals if two predictions are too close, where the lower confidence one is
ignored.} 
For each prediction, we use the position and the direction of the actor at
$t=0$ as well as those at the goal to interpolate a curve, using Bezier
quadratic parameterization. 
We then unroll a constant acceleration kinematic model along this curve, and sample 2D
points at each future timestep based on the curve and the kinematic information.
These 2D points form a trajectory, which serves as an initial proposal of our final forecasting.
Despite its simplicity, this parameterization gives us surprisingly good results.

Our final step is to refine those trajectory proposals using a learnable header. 
Similar to the shortcut layer introduced in Sec.
\ref{sec:backbone}, we use a lane pooling followed by a 1D CNN to 
pool features of this trajectory. Finally, we decode a pair of values per timestep, representing the residue from the trajectory proposal to the
ground-truth future position at this timestep (encoded in Frenet coordinate of
this trajectory proposal). We provide more detailed definitions of our
parameterization and output space in the supplementary~\ref{sec:supp_output}.

\subsection{Learning}
\label{sec:learning}
We train our model end-to-end with a loss containing the goal classification, the goal
regression, and the trajectory refinements. Specifically, we use
$$
\label{eq:objective}
\mathcal{L} = \mathcal{L}_{\text{cls}} + \alpha\mathcal{L}_{\text{reg}} +
\beta\mathcal{L}_{\text{refine}},
$$
where $\alpha$ and $\beta$ are hyparameters determining relative weights of
different terms. As our model predicts the goal classification and regression results per node, we simply adopt a binary cross entropy loss for $\mathcal{L}_{\text{cls}}$ with online hard example mining \cite{ohem} and a smooth-L1 loss for $\mathcal{L}_{\text{reg}}$, where
the $\mathcal{L}_{\text{reg}}$ is only evaluated on positive nodes, \ie closest lane
segments to the ground-truth final positions. The $\mathcal{L}_{\text{refine}}$ is also
a smooth-L1 loss with training labels generated on the fly: projecting
ground-truth future trajectories to the predicted trajectory proposals, and use
the Frenet coordinate values as our regression targets.

\begin{table*}[t]
\centering
\begin{tabular}{c|cc|cc|ccc}
  \specialrule{.2em}{.1em}{.1em}
  \multirow{2}{*}{Module} & \multicolumn{2}{c|}{\multirow{2}{*}{Ablation}} &
  \multicolumn{2}{c|}{K=1} & \multicolumn{3}{c}{K=6} \\
 & & & minADE & minFDE & minADE & minFDE & MR \\
  \hline
  \multirow{7}{*}{LaneRoI Encoder} & \ROI & Shortcut & & & & & \\
  \cline{2-8}
              & & & 1.68 & 3.79 & 0.86 & 1.46 & 14.5 \\
  & \checkmark & & 1.68 & 3.84 & 0.82 & 1.36 & 12.9 \\
  & \checkmark & Global Pool & 1.69 & 3.84 & 0.84 & 1.38 & 12.8 \\
  & \checkmark & Center Pool & 1.67 & 3.80 & 0.83 & 1.35 & 12.4 \\
  & \checkmark & Ours x 1 & 1.55 & 3.45 & 0.81 & \textbf{1.29} & 11.1 \\
  \rowcolor{grey} \cellcolor{white}& \checkmark & Ours x 2 & \textbf{1.54} &
  \textbf{3.45} & \textbf{0.80} & \textbf{1.29} & \textbf{10.8}\\
  \specialrule{.1em}{.05em}{.05em}
  \specialrule{.1em}{.05em}{.05em}
  \multirow{8}{*}{LaneRoI Interactor} & Interactor-Arch & Pooling & & & & & \\
  \cline{2-8}
                              & & & 1.54 & 3.45 & 0.80 & 1.29 & 10.8 \\
  & Attention & Global & 1.42 & 3.10 & 0.78 & 1.24 & 9.8 \\
  & Attention & Shortcut & 1.47 & 3.22 & 0.80 & 1.25 & 10.1 \\
  & GNN & Global & 1.45 & 3.15 & 0.79 & 1.25 & 9.9 \\
  & GNN & Shortcut & 1.45 & 3.21 & 0.79 & 1.25 & 10.0 \\
  & Ours & AvgPool & 1.42 & 3.11 & 0.79 & 1.25 & 9.9 \\
  \rowcolor{grey} \cellcolor{white} & Ours & LanePool & \textbf{1.33} &
  \textbf{2.85} & \textbf{0.77} & \textbf{1.19} & \textbf{8.2}\\
  \specialrule{.1em}{.05em}{.05em}

\end{tabular}
\caption{Ablations on different modules of LaneRCNN. Metrics are reported on the validation set.
In the upper half, we examine our \ROI Encoder, comparing per-actor 1D feature vector 
v.s. \ROI representations as well as different designs for the shortcut mechanism.
In the lower half, we compare different strategies to model interactions, including a
fully connected graph among actors with GNN / attention, as well as ours.
Pooling refers to how we pool a 1D actor feature from each \ROI which are
used by GNN / attention. Rows shaded in gray indicate the architecture used in
our final model.}
\label{table:ablation}
\end{table*}

\section{Experiment}

We evaluate the effectiveness of LaneRCNN on the large-scale Argoverse motion
forecasting benchmark (Argoverse), which is publicly available and provides annotations
of both actor motions and HD maps. In the following, we first
explain our experimental setup and then compare our method against
state-of-the-art on the leaderboard. We also conduct ablation studies on each
module of LaneRCNN to validate our design choices. Finally, we present some
qualitative results.

\subsection{Experimental Settings}
\paragraph{Dataset:}
Argoverse provides a large-scale dataset \cite{argoverse} for the
purpose of training, validating and testing models, where the task is to
forecast 3 seconds future motions given 2 seconds past observations. 
This dataset consists of more than 30K real-world driving sequences collected in Miami and Pittsburgh.
Those sequences are further split into train, validation, and test sets without
any geographical overlapping. Each of them has 205942, 39472, and 78143 sequences
respectively. In particular, each sequence contains the positions of all actors in
a scene within the past 2 seconds history, annotated at 10Hz.
It also specifies one interesting actor in this scene, with type `agent', whose future 3 seconds
motions are used for the evaluation. The train and validation splits additionally
provide future locations of all actors within 3 second horizon labeled at 10Hz, while annotations for test sequences are withheld from the public and used for
the leaderboard evaluation. Besides, HD map information can be retrieved for all
sequences. 

\paragraph{Metrics:} We follow the benchmark setting and use Miss-Rate (MR), Average
Displacement Error (ADE) and Final Displacement Error (FDE), which are also
widely used in the community. MR is defined as the ratio of data that none of
the predictions has less than 2.0 meters L2 error at the final timestep. ADE is
the averaged L2 errors of all future timesteps, while FDE only
counts the final timestep. To evaluate the mutli-modal prediction, we
also adopt the benchmark setting: predicting K=6 future trajectories
per actor and evaluating the $\text{min}_{K}\text{MR}$, $\text{min}_{K}\text{ADE}$,
$\text{min}_{K}\text{FDE}$ using the trajectory that is closest to the
ground-truth.

\paragraph{Implementation Details:}
We train our model on the \emph{train} set with the batch size of 64 and terminate at 30
epochs. We use Adam \cite{adam} optimizer with the learning rate initialized at 0.01 and decayed by 10 at 20 epochs. To normalize the data, we translate and rotate the coordinate system of each sequence so that
the origin is at current position ($t=0$) of `agent' actor and x-axis is aligned
with its current direction. During training, we further apply a random rotation
data augmentation within $(-\frac{2}{3}\pi, \frac{2}{3}\pi)$. No other data
processing is applied such as label smoothing. More implementation
details are provided in the supplementary~\ref{sec:supp_implement}.

\subsection{Comparison with State-of-the-art}
We compare our approach with top entries on Argoverse motion forecasting
leaderboard \cite{argoleaderboard} as well as official baselines provided
by the dataset \cite{argoverse} as shown in Table \ref{table:argo}. 
We only submit our final model once to the leaderboard and achieve
state-of-the-art performance.\footnote{Snapshot of the leaderboard at the submission time: Nov. 12, 2020.}
This is a very challenging benchmark
with around 100 participants at the time of our submission. Note that for the
official ranking metric MR (K=6), previous leading methods are extremely close to
each other, implying the difficulty of further improving the performance.
Nevertheless, we significantly boost the number which verifies the
effectiveness of our method. Among the competitors, both Jean
\cite{mercat2020multi} and TNT \cite{tnt} use RNNs to encode actor kinematic
states and lane polylines. They then build a fully-connected interaction graph
among all actors and lanes, and use either the attention or GNNs to model
interactions. As a result, they represent each actor with a single
feature vector, which is less expressive than our \ROI representations. Moreover,
the fully-connected interaction graph may also discard valuable map structure
information. Note that TNT shares a similar output parameterization as
ours, yet we perform better on all metrics. This further validates the advantages of
our \ROI compared against traditional representations. Unfortunately, since Poly team does
not publish their method, we can not compare with it qualitatively.

\begin{figure*}[t]
\centering
\setlength{\tabcolsep}{1pt}
\begin{tabular}{cccc}
\includegraphics[width=0.24\linewidth]{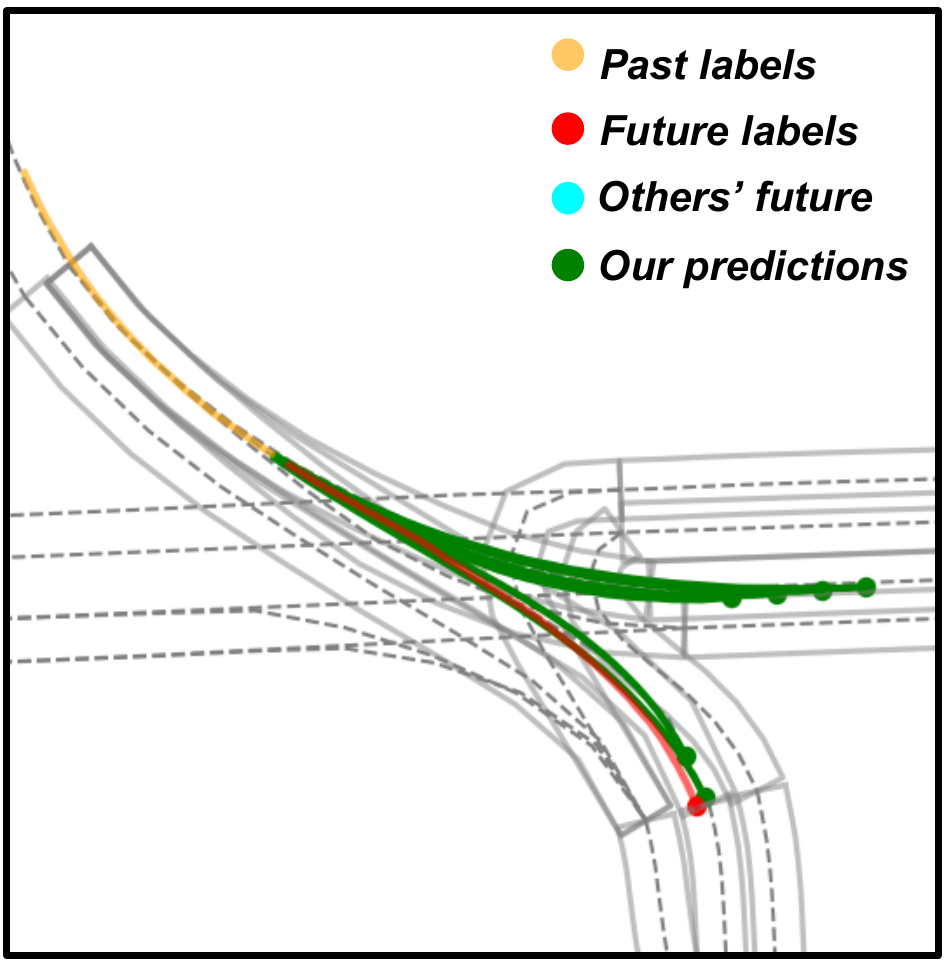}
&\includegraphics[width=0.24\linewidth]{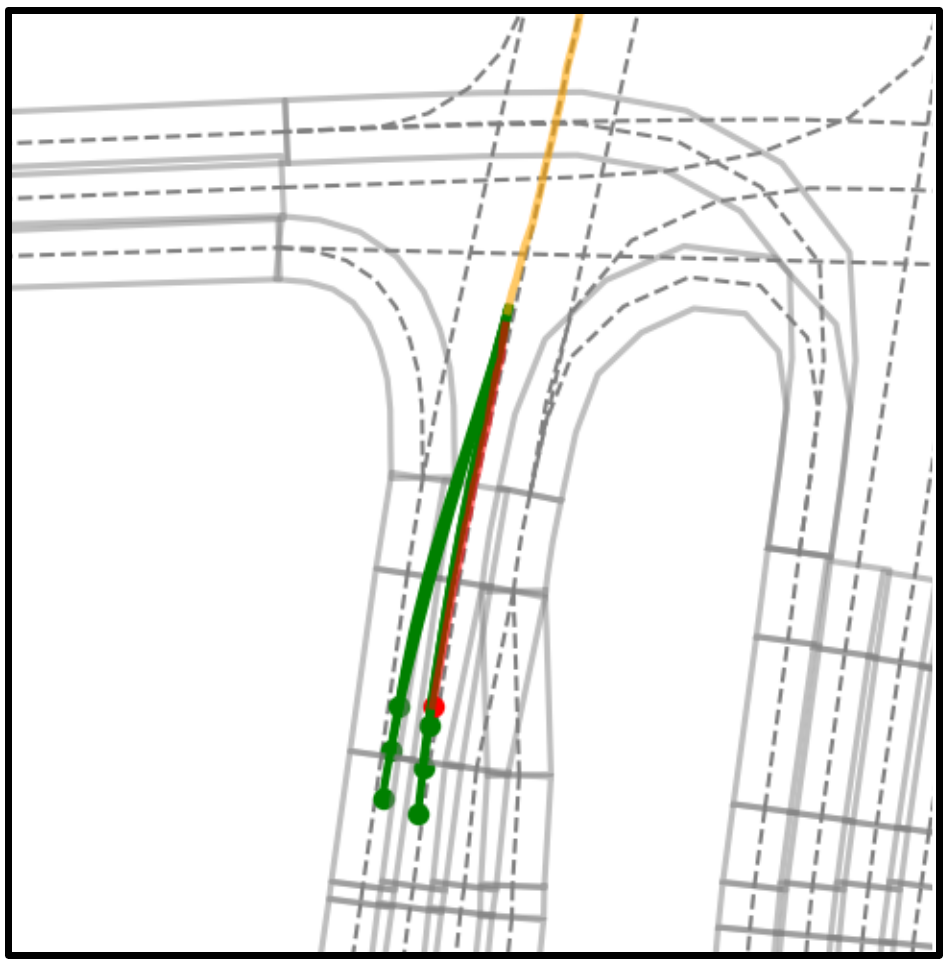}
&\includegraphics[width=0.24\linewidth]{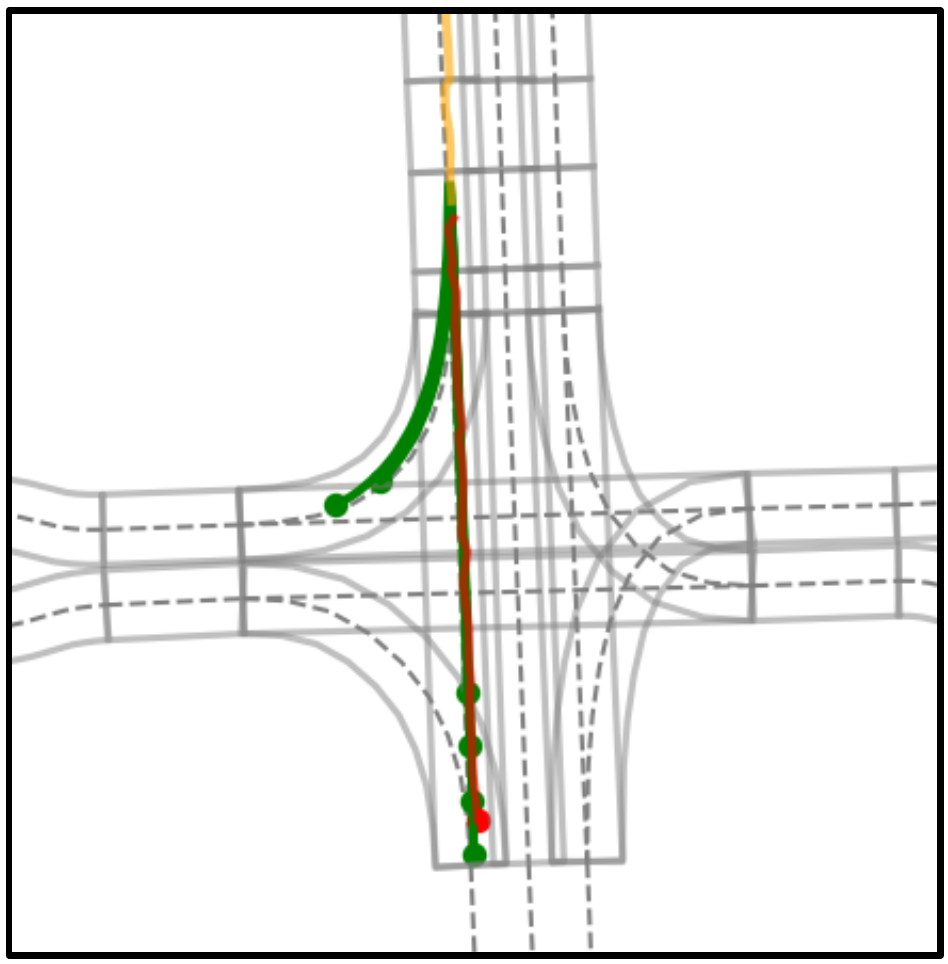}
&\includegraphics[width=0.24\linewidth]{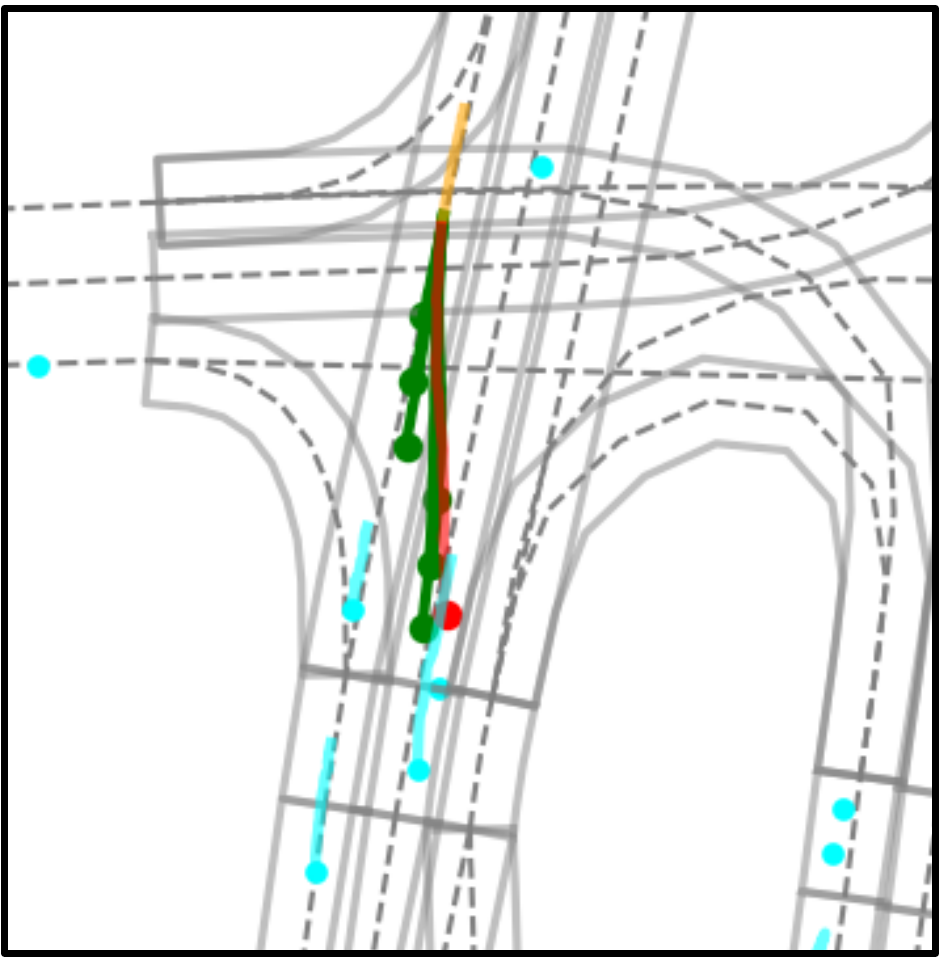}

\end{tabular}
\caption{Qualitative results on Argoverse validation set. Here we show (from
left-to-right): 1) curved lanes 2) lane changing 3) intersection 4) overtaking.}
\label{fig:vis}
\end{figure*}

\subsection{Ablation Studies}

\paragraph{Ablations on LaneRoI Encoder:}
We first show the ablation study on one of our main contributions, \ie, \ROI, in the upper half of Table
\ref{table:ablation}. 
The first row shows a representative of the traditional representations. 
Specifically, we first build
embeddings for lane graph nodes using only the map information and 4 lane
convolution layers. We then use a
1D CNN (U-net style) to extract a motion feature vector from actor kinematic states, concatenate it
with every graph node embedding and make predictions. Conceptually, this is
similar to TNT \cite{tnt} except that we modify the backbone network to make comparisons fair. 
On the second row, we show the result of our \ROI
representations with again four lane convolution layers (no shortcuts). Hence, the only difference
is whether the actor is encoded with a single motion vector shared by
all nodes, or encoded in a distributed and structured manner as ours. As shown
in the table, our \ROI achieves similar or better results on all
metrics, exhibiting its advantages. Note that this row is not yet our best result
in terms of using \ROI representations, as the actor information is only exposed
to a small region during the input encoding (clamping at input
node embeddings) and can not efficiently propagate to
the full \ROI without the help of the shortcut, which we will show next.

Subsequent rows in Table \ref{table:ablation} compare different design
choices for the shortcut mechanism, in particular how we pool the global feature
for each \textit{LaneRoI}. `Global Pool' refers to average-pooling all node
embeddings within a \textit{LaneRoI}, and `Center Pool' means we pool a feature
from a \ROI using nodes that around the last observation of the actor and a lane
pooling. As we can see, although these two approaches can possibly
spread out information to every node in a \textit{LaneRoI} (and thus build a
shortcut), they barely improve the performance. On the contrary, ours achieve
significant improvements. This is because we pool features along the past
trajectory of the actor, which results in a larger and actor-motion-specific receptive field.
Here, $\times 1$ and $\times 2$ refer to an encoder with 1 shortcut per 4 and 2 lane
convolution layers respectively. This shows stacking more shortcuts
provides some, but diminishing, benefits.

\paragraph{Ablations on LaneRoI Interactor:}
To verify the effectiveness of our map-aware interaction module, we compare against several model variants based
on the fully-connected interaction graph among actors. Specifically, for each actor,
we apply a \ROI encoder\footnote{We choose \ROI encoder rather than other
encoder, \eg, CNN, for fair comparisons with ours.} to process node embeddings, and then pool an
actor-specific feature vector from \ROI via either the global average pooling or our
shortcut mechanism. These actor features are then fed into a transformer-style
\cite{transformer} attention module or a fully-connected GNN.
Finally, we add the output actor features to nodes in their \ROI respectively
and make predictions using our decoding module.
As a result, these variants have the same pipeline as ours, with the only
difference on how to communicate across actors. To make comparisons
as fair as possible, both the attention and GNN have the same
numbers of layers and channels as our \ROI Interactor.\footnote{The GNN here is almost identical to our lane
convolution used in Interactor except for removing the multi-hop as the graph is fully-connected.}

As shown in Table \ref{table:ablation}, all interaction-based models outperform
the one without considering interactions (row 1) as expected. In addition, our
approach significantly improves the performance compared to both the attention and GNN. 
Interestingly, all fully-connected interaction graph based model reach similar
performance, which might imply such backbones may saturate the performance (as
also shown by leading methods on the leaderboard).
We also show that naively using the average pooling to embed features from
\textit{LaneRoI}s to global graph does not achieves good performance
because it ignores local structures.

\subsection{Qualitative results}
In Figure~\ref{fig:vis}, we show some qualitative results on Argoverse validation
set. We can see that our method generally follows the map very well and demonstrates good
multi-modalities. From left to right, we show 1) when an actor follows a
curved lane, our model predicts two direction modes with different velocities;
2) when it is on a straight lane, our model covers the possibilities of lane changing; 3)
when it's approaching an intersection, our model captures both the go-straight and the
turn-right modes, especially with lower speeds for turning right, which are
quite common in the real world; 4) when there is an actor blocking the path, we
predict overtaking behaviors matching exactly with the ground-truth. Moreover,
for the lane-following mode, we predict much slower speeds which are consistent with
this scenario, showing the effectiveness of our interaction modeling. 
We provide more qualitative results in the supplementary~\ref{sec:supp_qual}.

\section{Conclusion}

In this paper, we propose LaneRCNN, a graph-centric motion forecasting model.
Relying on learnable graph operators, LaneRCNN builds a distributed lane-graph-based
representation (\textit{LaneROI}) per actor to encode its past motion and the local map topology.
Moreover, we propose an interaction module which effectively captures the interactions among actors within the shared global lane graph.
And lastly, we parameterize the output trajectory using lane graphs which helps improve the prediction.
We demonstrate that LaneRCNN achieves state-of-the-art performance on the challenging Argoverse motion forecasting benchmark.

\section*{Acknowledgement}
We would like to sincerely thank Siva Manivasagam, Yun Chen, Bin Yang, Wei-Chiu Ma and
Shenlong Wang for their valuable help on this paper.

{\small
\bibliographystyle{ieee_fullname}
\bibliography{mybib}
}

\clearpage
\appendix

\section{Map-Relative Output Decoding}
\label{sec:supp_output}
Our output decoding can be divided into three steps: predicting the final goal
of an actor based on node embeddings, propose an initial trajectory based on the
goal and the initial pose, and refine the trajectory proposal with a learnable
header. As the first step is straight-forward, we now explain how we perform the
second and third steps in details.

\subsection{Trajectory Proposal}
Given a predicted final pose $\left(x^T, y^T, dx^T, dy^T\right)$ and initial
pose $\left(x^0,
y^0, dx^0, dy^0\right)$ of an actor, where $(x, y)$ is the 2D location and $(dx, dy)$
is the tangent vector, we fit a Bezier quadratic curve satisfying these boundary
conditions, \ie, zero-th and first order derivative values. Specifically, the curve can be parameterized by 
\begin{align}
  \nonumber
  x(s) &= a_0 s^2 + a_1 s + a_2,
  \\
  \nonumber
  y(s) &= b_0 s^2 + b_1 s + b_2,
  \\
  \nonumber
  s.t. \quad x(0) &= x^0, \quad x(1) = x^T, \quad \frac{x'(0)}{x'(1)} = \frac{dx^0}{dx^T},
  \\
  \nonumber
  y(0) &= y^0, \quad y(1) = y^T,  \quad \frac{y'(0)}{y'(1)} =
  \frac{dy^0}{dy^T}.
\end{align}
Here, $s$ is the normalized distance.
As a result, each predicted goal uniquely defines a 2D curve.

Next, we unroll a velocity profile along this curve to get 2D waypoint proposals
at every future timestamp. Assuming the actor is moving with a constant
acceleration within the prediction horizon, we can compute the acceleration
based on the initial velocity $v$ and the traveled distance $s$ (from
$(x^0, y^0)$ to $(x^T, y^T)$ along the Bezier curve) using
$$
a = \frac{2\times(s - vT)}{T^2}.
$$
Therefore, the future position of the actor at any timestamp $t$ can be
evaluated by querying the position along the curve at $s(t) = vt +
\frac{1}{2}at^2$.

\subsection{Trajectory Refinement}
Simply using our trajectory proposals for motion forecasting
will not be very accurate, as not all actors
move with constant accelerations and follow Bezier curves, yet the proposals
provide us good initial estimations which we can further refine. To do
so, for each trajectory proposal, we construct its features using a shortcut
layer on top our \ROI node embeddings. We then use a 2 layer MLP to decode a
pair of values $\left(s^t, d^t\right)$ for each future timestamp, representing
the actor position at time $t$ in the Frenet Coordinate System~\cite{frenet}.
The Cartesian coordinates $\left(x^t, y^t\right)$ can be mapped from
$\left(s^t, d^t\right)$ by first traversing along the Bezier curve distance
$s^t$ (\aka longitudinal), and then deviating perpendicularly from the curve
distance $d^t$ (\aka lateral). The sign of $d^t$ indicates the deviation is
either to-the-left or to-the-right.

\begin{table*}[t]
  \centering
  \begin{tabular}{ccc|ccccc}
  \specialrule{.2em}{.1em}{.1em}
  Sampling & Avg Length ($\ell_k$) & Reg & $\text{min}_1$ADE & $\text{min}_1$FDE & $\text{min}_6$ADE & $\text{min}_6$FDE & $\text{min}_6$MR\\
  \hline
  up 1x & 2.1 m& & 1.52 & 3.32 & 0.94 & 1.69 & 24.0\\
  up 1x & 2.1 m& \checkmark & 1.41 & 3.03 & 0.83 & 1.35 & 14.1\\
  up 2x & 1.1 m& & 1.43 & 3.09 & 0.85 & 1.39 & 13.4\\
  up 2x & 1.1 m& \checkmark & 1.39 & 2.99 & 0.80 & 1.24 & 10.2\\
  uniform & 2 m & & 1.39 & 2.94 & 0.86 & 1.44 & 10.5 \\
  uniform & 2 m & \checkmark & 1.35 & 2.86 & 0.80 & 1.29 & \textbf{8.2}\\
  uniform & 1 m & & 1.37 & 2.90 & 0.83 & 1.32 & 9.9\\
  \rowcolor{grey}uniform & 1 m & \checkmark & \textbf{1.33} & \textbf{2.85} &
  \textbf{0.77} & \textbf{1.19} & \textbf{8.2}\\
  \specialrule{.1em}{.05em}{.05em}

\end{tabular}
\caption{Ablation studies on lane segments sampling strategy and regression
  branch. We compare
  different sampling strategies including using the original segment labels (upsample 1x) or
  upsample the labels (upsample 2x), as well as uniformly sample segments along
  lanes (uniform 2/1m). For each, we also compare using the regression branch or
not (only classification branch). Our final model is shaded in \colorbox{grey}{grey}.}
\label{table:supp_lane}
\end{table*}

\section{LaneRoI Construction}
\label{sec:supp_roi}
To construct a \textit{LaneRoI}, we need to retrieve all relevant lanes of a
given actor. In our experiments, we use a simple heuristic for this purpose.
Given an HD map of a scene and an arbitrary actor in this scene, we first
uniformly sample segments $\ell_k$ with 1 meter length along each lane's centerline
. Then, for the actor's location at each past timestamp, we find the nearest
lane segment and collect them together into a set.
This is a simplified version
of lane association and achieve very high recall of the true ego-lane.
Finally, we retrieve predecessing and successing lane segments (within a range
$D$) of those segments in the set to capture lane-following behaviors, 
as well as left and right neighbors
of those predecessing and successing lanes which are necessary to capture lane-changing behaviors.
The range $D$ equals to an expected length of future movement by integrating
the current velocity within the prediction horizon, plus a buffer value, \eg, 20
meters. Therefore, $D$ is dynamically changed based on actor velocity. This is
motivated by the fact that high speed actors travel larger distances and thus
should have larger \textit{LaneRoI}s to capture their motions as well as interactions with
other actors.

\section{Architecture and Learning Details}
\label{sec:supp_implement}
Our LaneRCNN is constructed as follows: we first feed each input \ROI
representation into an encoder, consists of 2 lane convolution layers
and a shortcut layer, followed by another 2 lane convolution layers and a
shortcut layer. We then use a lane pooling layer to build the node embeddings of
the global graph (interactor), where the neighborhood threshold is set to 2
meters. Another four layers of lane convolution are applied on top of this global
graph. Next, we distribute the global node embeddings to each \ROI by
using a lane pooling layer and adding the pooled features to original \ROI
embeddings (previous encoder outputs) as a skip-connection architecture. Another
4 layers of lane convolution and 2 layers of shortcut layers are applied
afterwards. Finally, we use two layers of MLP to decode a classification score
per node using its embeddings, and another two layer for regression branch as
well. All layers except the final outputs have 64 channels. We use Layer
Normalization~\cite{layernorm} and ReLU~\cite{relu} for normalization and
non-linearity respectively.

During training, we apply online hard example mining~\cite{ohem} for
$\mathcal{L}_{cls}$ (Eq.~\ref{eq:objective}). Recall each node predict a binary classification score
indicating whether this node is the closest lane segment to the final actor
position. We use the closest lane segment to the ground-truth location as our
positive example. Negative examples are all nodes deviating from the
ground-truth by more than 6 meters and the remaining nodes are `don't care' which do
not contribute any loss for $\mathcal{L}_{cls}$. Then, we randomly subsample one
fourth of all negative examples and among them we use the hardest 100 negative
examples for each data sample to compute our loss. The final $\mathcal{L}_{cls}$
is the average of positive example loss plus the average of negative example
loss. Finally, we add $\mathcal{L}_{reg}$ and $\mathcal{L}_{refine}$ with
relative weights of $\alpha = 0.5$ and $\beta = 0.2$ respectively to form our
total loss $\mathcal{L}$.

\section{Ablation Studies}
\label{sec:supp_ablation}
When constructing our \textit{LaneRoI}, we define a lane segment $\ell$ to be a node in the
graph. However, there are different ways to sample lane segments and we find
such a choice largely affects the final performance. In
Table~\ref{table:supp_lane} we show different strategies of sampling lane segments.
The first four rows refer to upsampling the original lane segment labels
provided in the dataset.\footnote{Lanes are labeled in the format of polylines
in Argoverse, thus points on those polylines naturally divide lanes into
segments.} Such a strategy provides segments with different lengths, \eg, shorter
and denser segments where the geometry of the lanes changes more rapidly. The
last four rows sample segments uniformly along lanes with a predefined length.

As we can observe from Table~\ref{table:supp_lane}, even though the `upsample' strategy
can result in similar average segment length as the `uniform' strategy, it
performs much worse in all metrics. This is possible because different lengths of
segments introduce additional variance and harm the representation learning
process. We can also conclude from the table that using denser sampling can
achieve better results. Besides, we show the effectiveness of adding a
regression branch for each node in addition to a classification branch, shown in
`Reg' column. 

\begin{figure}[t]
\begin{center}
  \includegraphics[height=6.6cm]{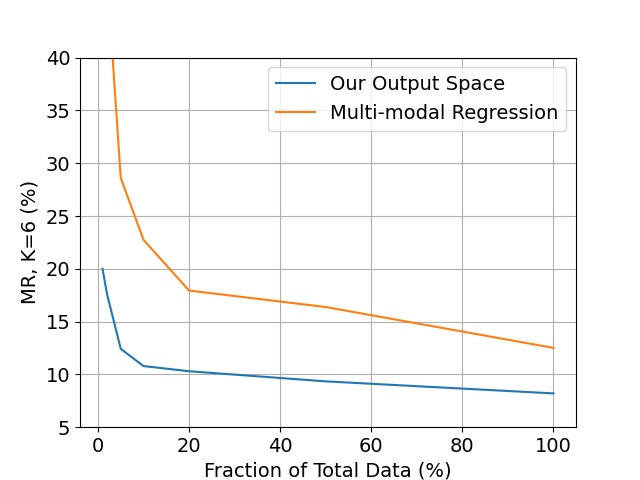}
\end{center}
\caption{Model performance when using different amounts of data for training.
  Our output parameterization explicitly leverages lanes as priors for motion
  forecasting, and thus significantly ease the learning compared to directly
  regressing the future motions in the 2D plane.
}
\label{fig:supp_output}
\end{figure}

Our output parameterization
explicitly leverages the lane information and thus ease the training. In
Fig.~\ref{fig:supp_output}, we validate such an argument where we compare against a
regression-based variant of our model. In particular, we use the same backbone
as ours, and then perform a shortcut layer on top of each \ROI to extract an
actor-specific feature vector. We then build a multi-modal regression header
which directly regresses future motions in the 2D Cartesian
coordinates.\footnote{As building such a header is non-trivial, we borrow an
open-sourced implementation in LaneGCN~\cite{lgn} which has tuned on the same
dataset and shown strong results.} We can see from Fig.~\ref{fig:supp_output} that our
model achieves decent performance when only small amounts of training
data are available: with only 1\% of training data, our method can achieve 20\%
of miss-rate. On the contrary, the regression-based model requires much more
data. This shows our method can exploit map structures as good priors and ease
learning.

In Table~\ref{table:supp_refinement}, we summarize ablations on different trajectory
parameterizations. We can see a constant acceleration rollout slightly improves
over constant speed assumption, and the Bezier curve significantly outperforms a
straight line parameterization, indicating it is more realistic. In addition,
adding a learnable header to refine the trajectory proposals (\eg, Bezier curve)
can further boost performance. 

\begin{table}[t]
  \centering
  \begin{tabular}{ccc|cc}
  \specialrule{.2em}{.1em}{.1em}
  Curve & Velocity & Learnable & $\text{min}_1$ADE & $\text{min}_6$ADE\\
  \hline
  line & const & & 1.53 & 1.04\\
  line & acc & & 1.52 & 1.02\\
  line & acc & \checkmark & 1.41 & 0.86\\
  Bezier & const & & 1.46 & 0.96\\
  Bezier & acc & & 1.44 & 0.94\\
  \rowcolor{grey}Bezier & acc & \checkmark & \textbf{1.33} & \textbf{0.77}\\
  \specialrule{.1em}{.05em}{.05em}

\end{tabular}
\caption{Ablation studies on output parameterizations. We compare different ways
of proposing curves (straight line v.s. Bezier quadratic curve), unrolling
velocities (const as constant velocity and acc as constant acceleration), as well
as using learnable refinement header or not. Our final model is shaded in
\colorbox{grey}{grey}.}
\label{table:supp_refinement}
\end{table}

\begin{figure*}[t]
\centering
\setlength{\tabcolsep}{1pt}
\begin{tabular}{cccc}
  \includegraphics[width=0.24\linewidth]{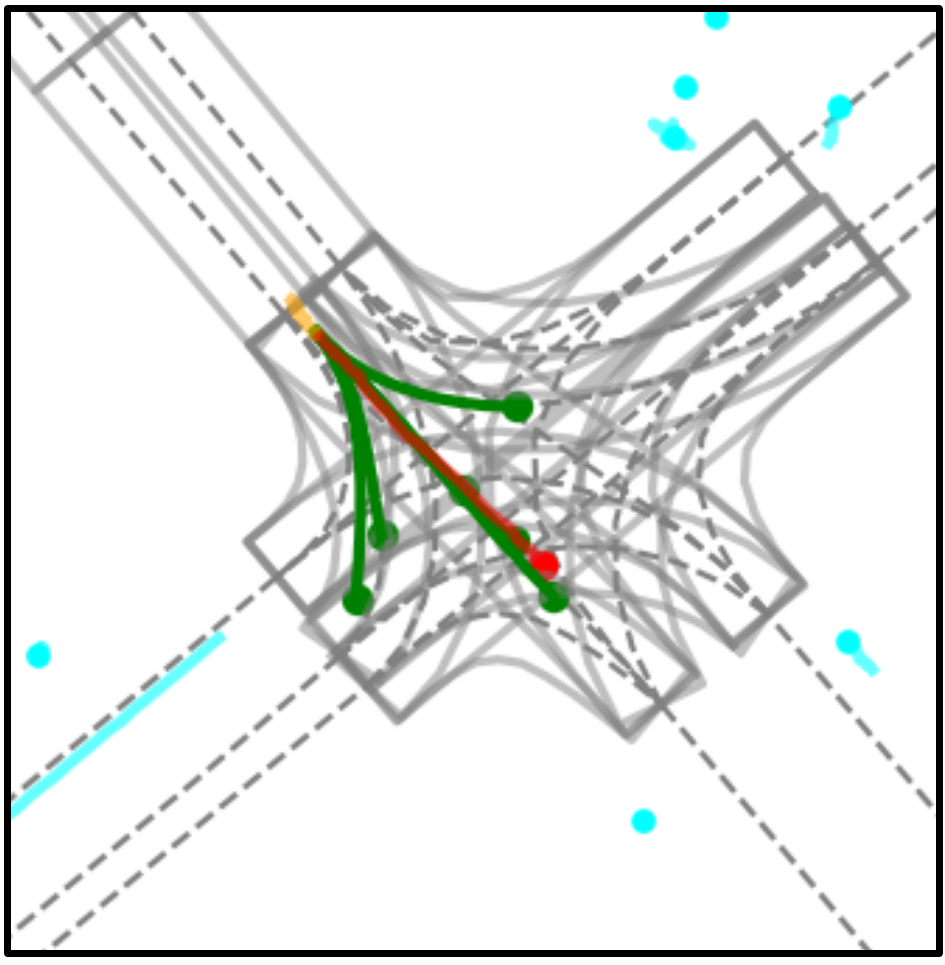}
&\includegraphics[width=0.24\linewidth]{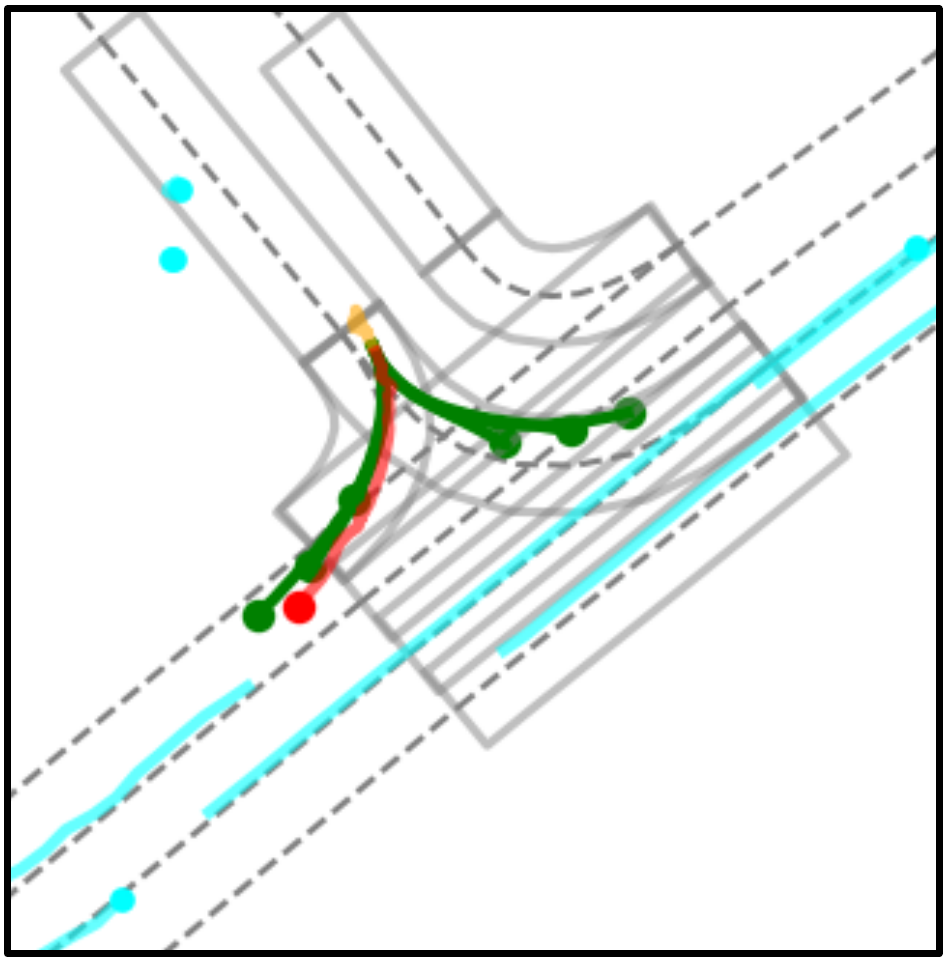}
&\includegraphics[width=0.24\linewidth]{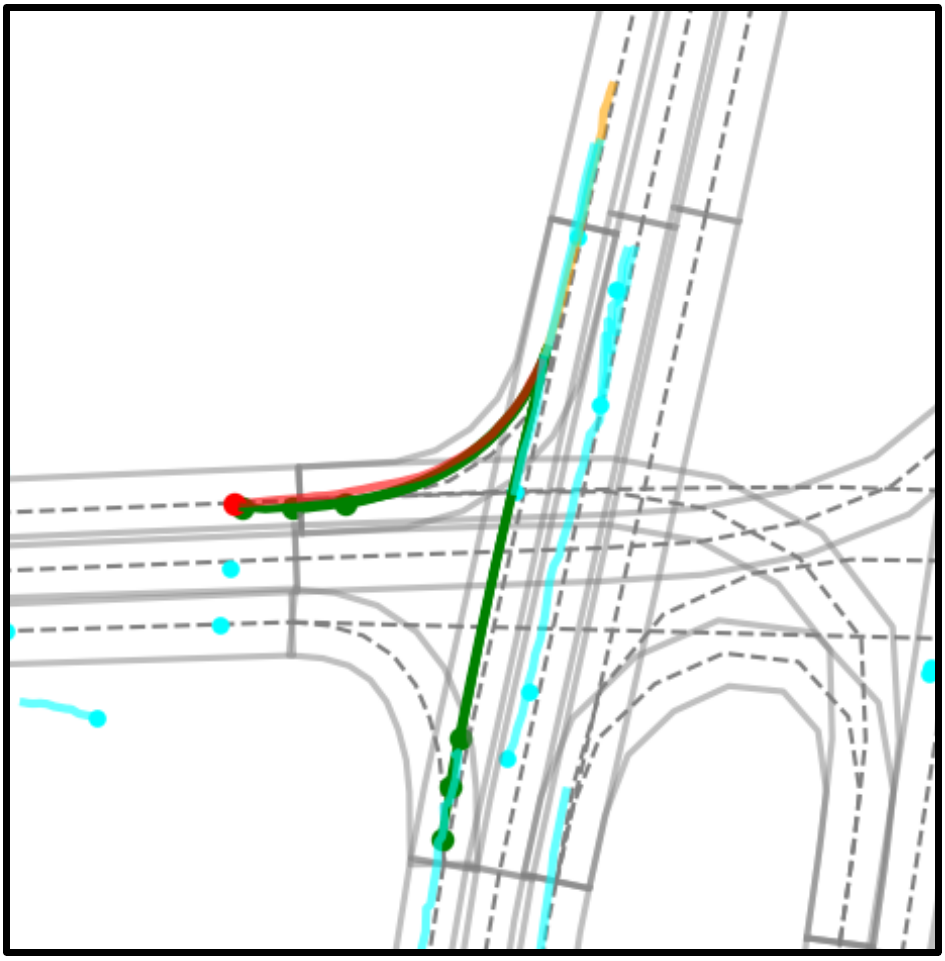}
&\includegraphics[width=0.24\linewidth]{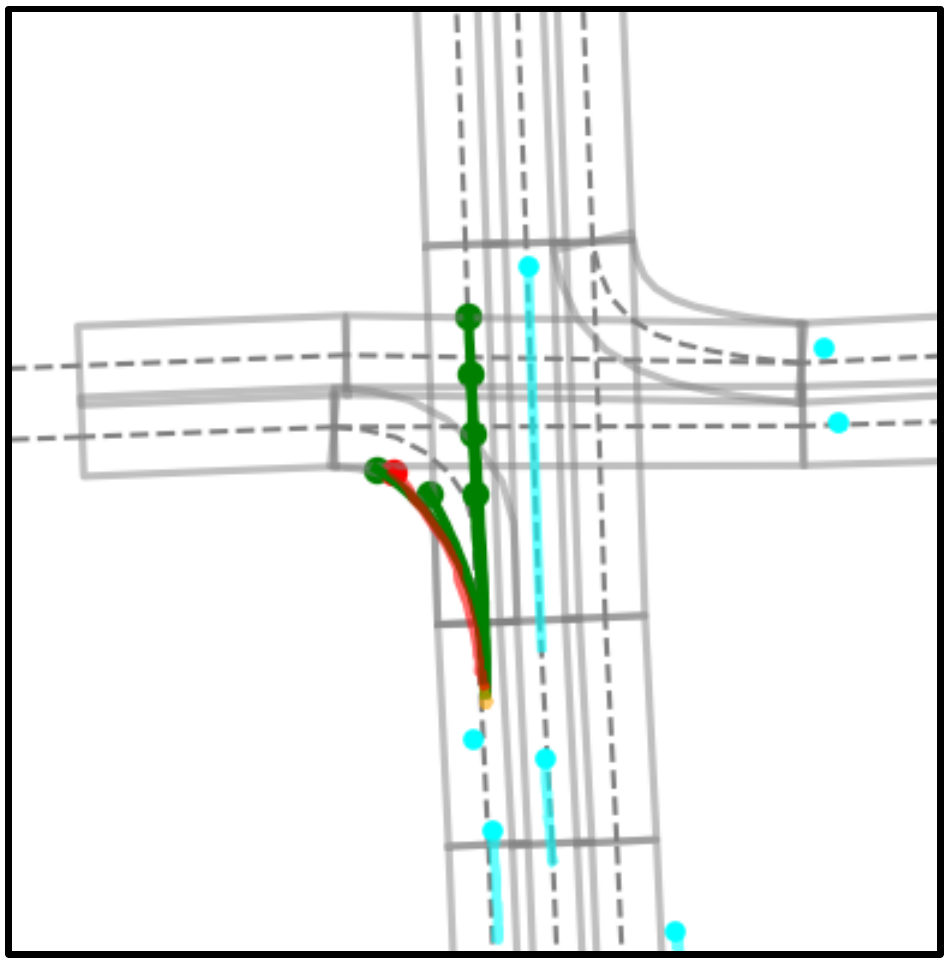}\\
\includegraphics[width=0.24\linewidth]{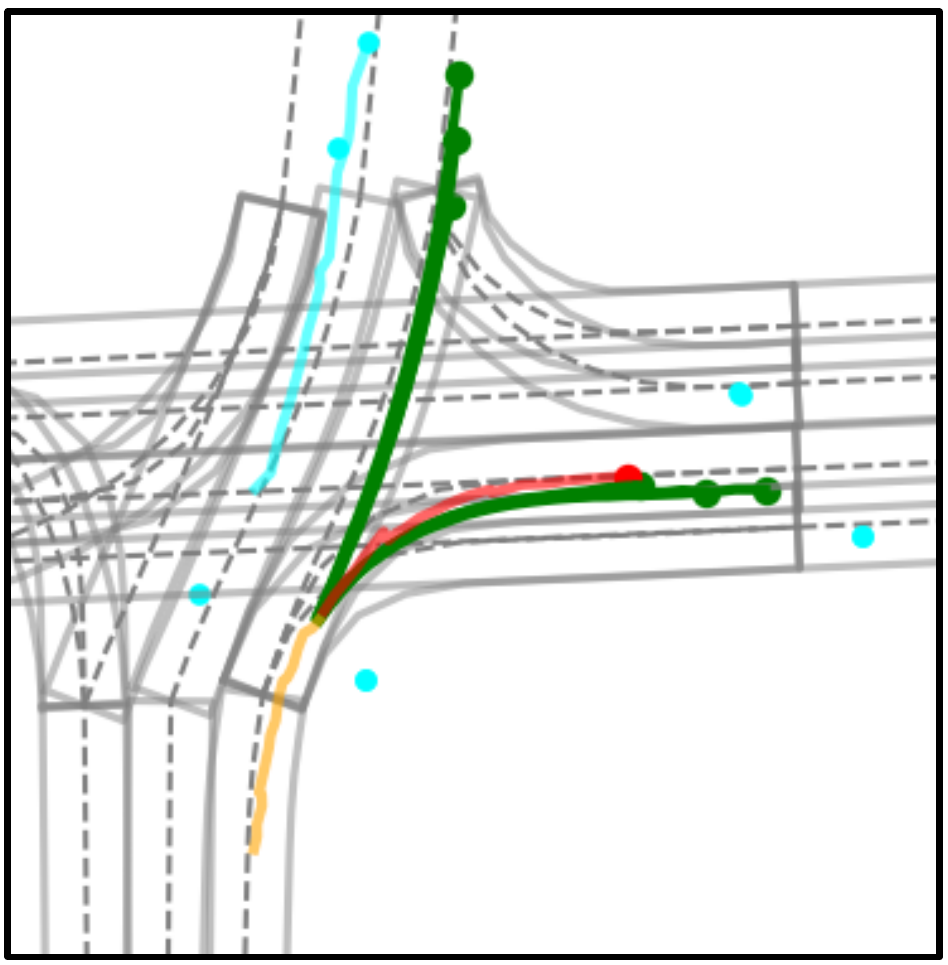}
&\includegraphics[width=0.24\linewidth]{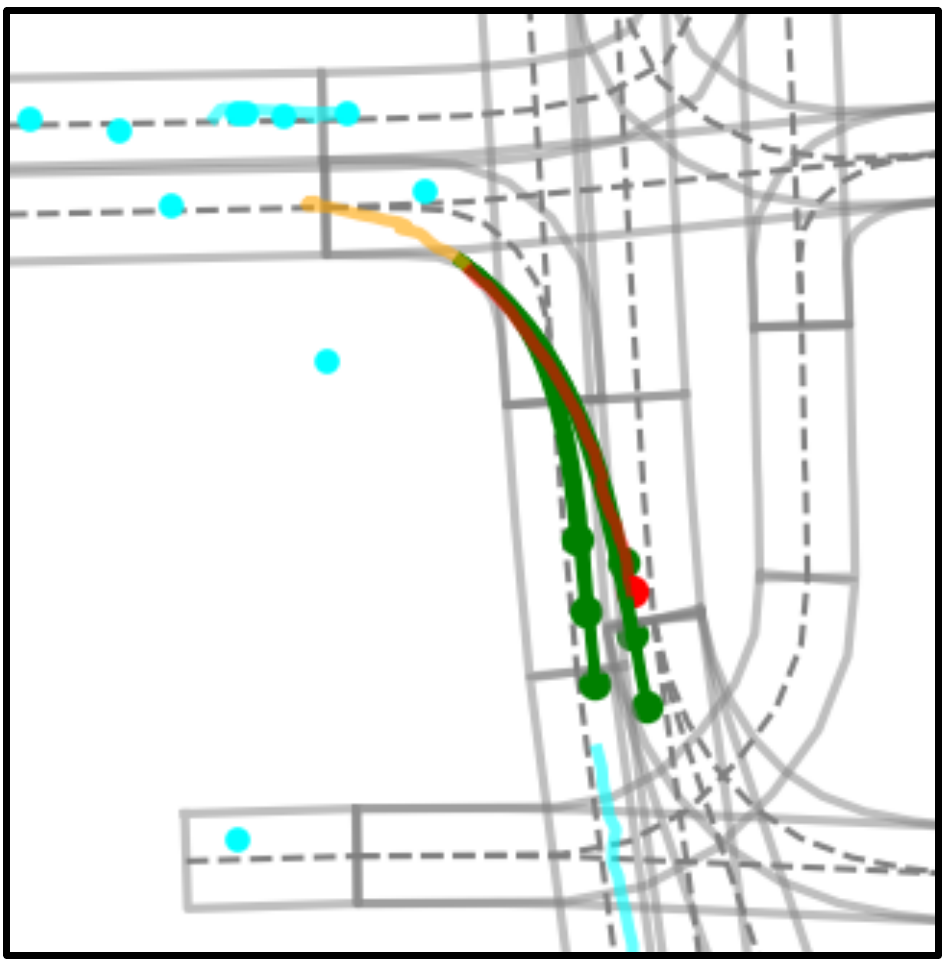}
&\includegraphics[width=0.24\linewidth]{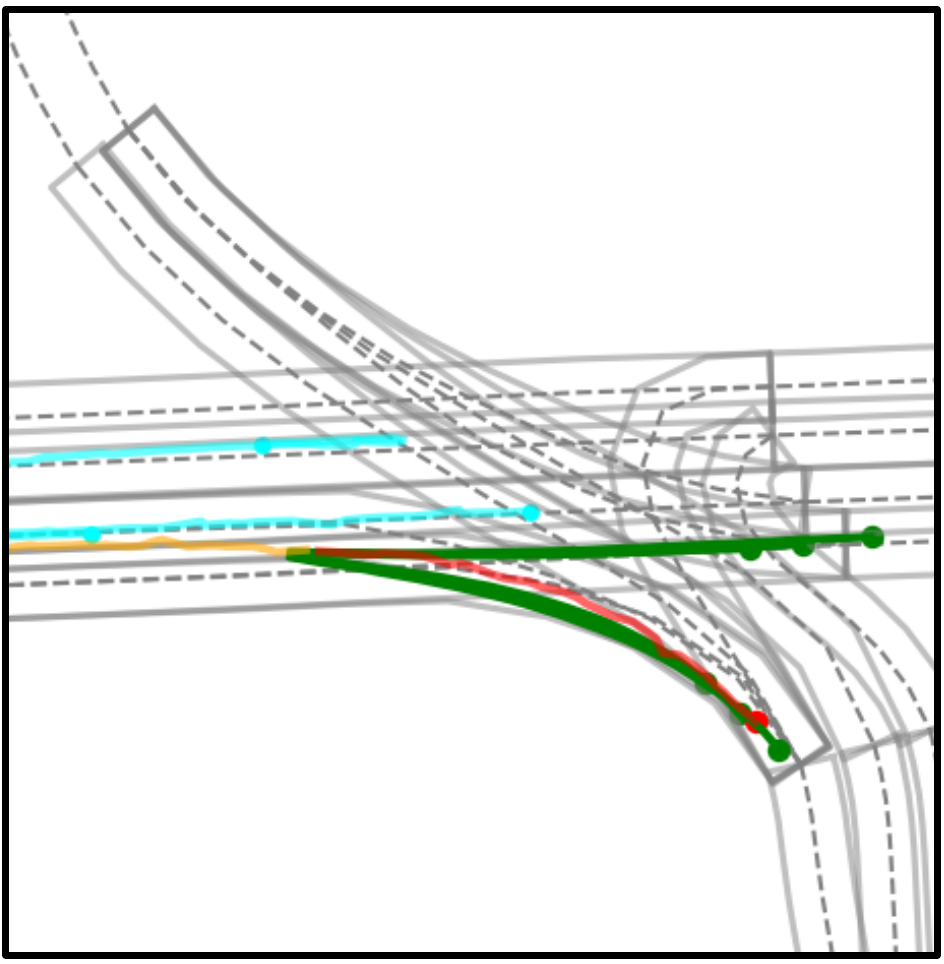}
&\includegraphics[width=0.24\linewidth]{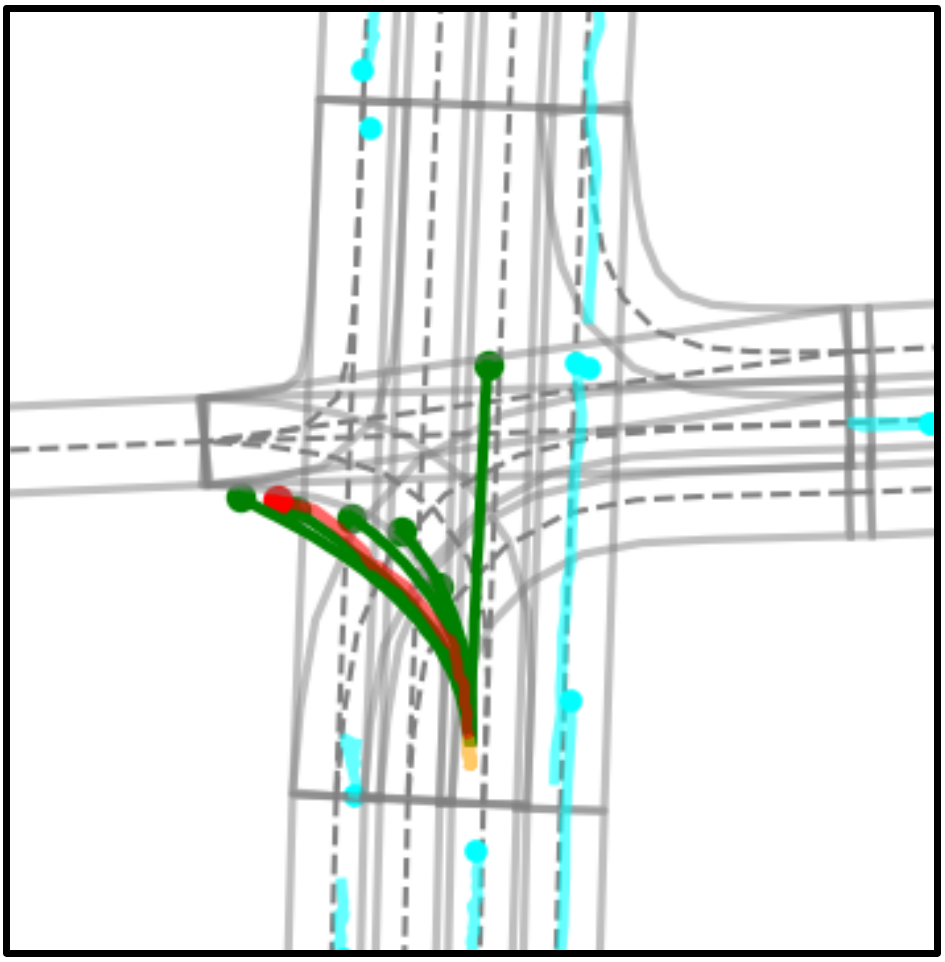}\\
\includegraphics[width=0.24\linewidth]{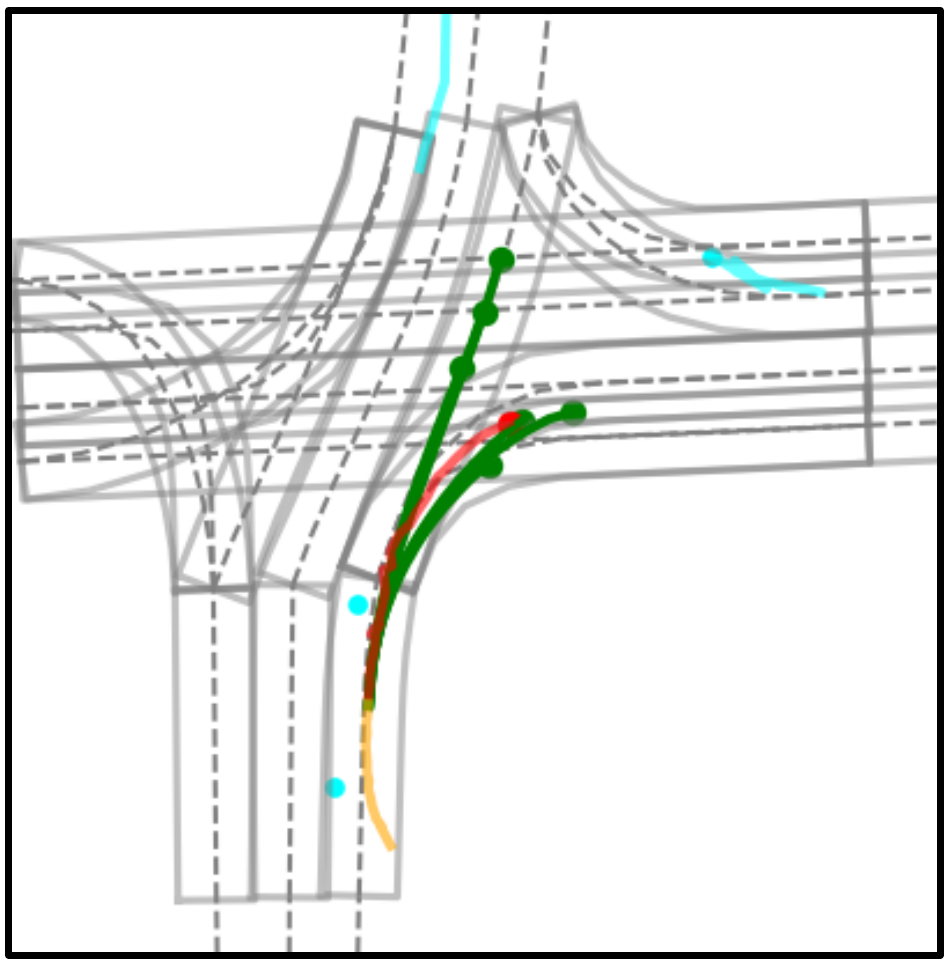}
&\includegraphics[width=0.24\linewidth]{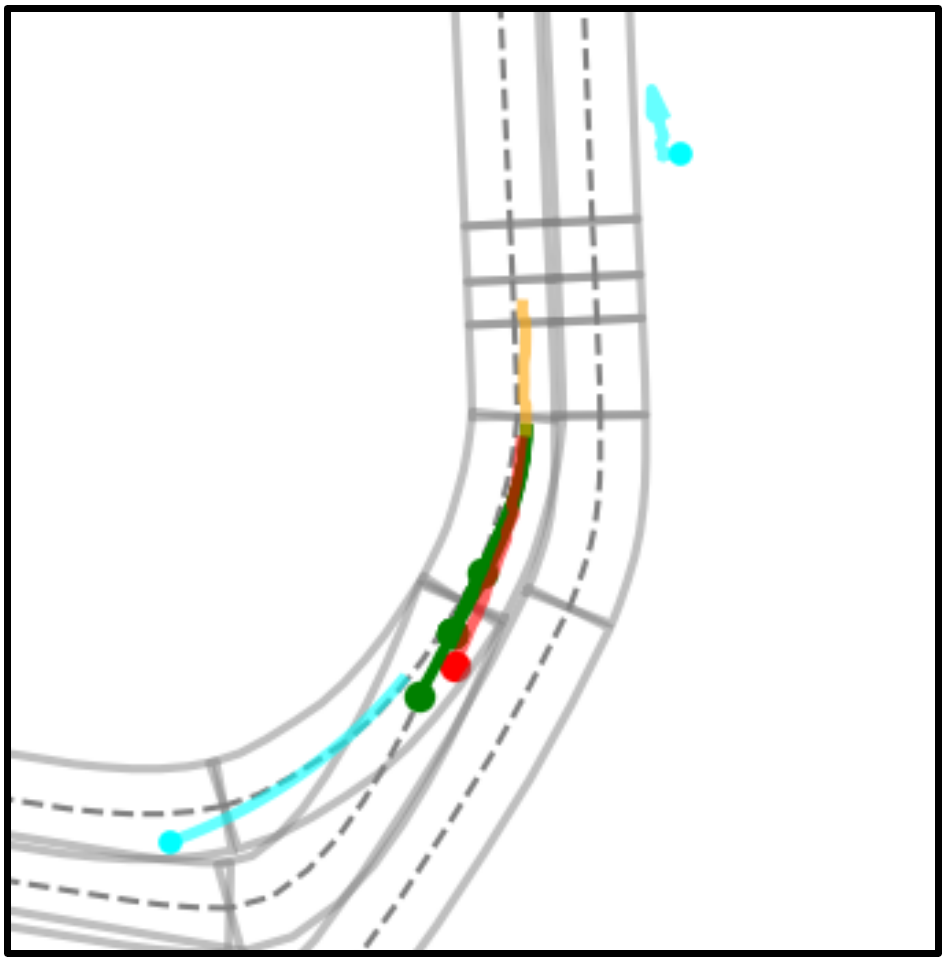}
&\includegraphics[width=0.24\linewidth]{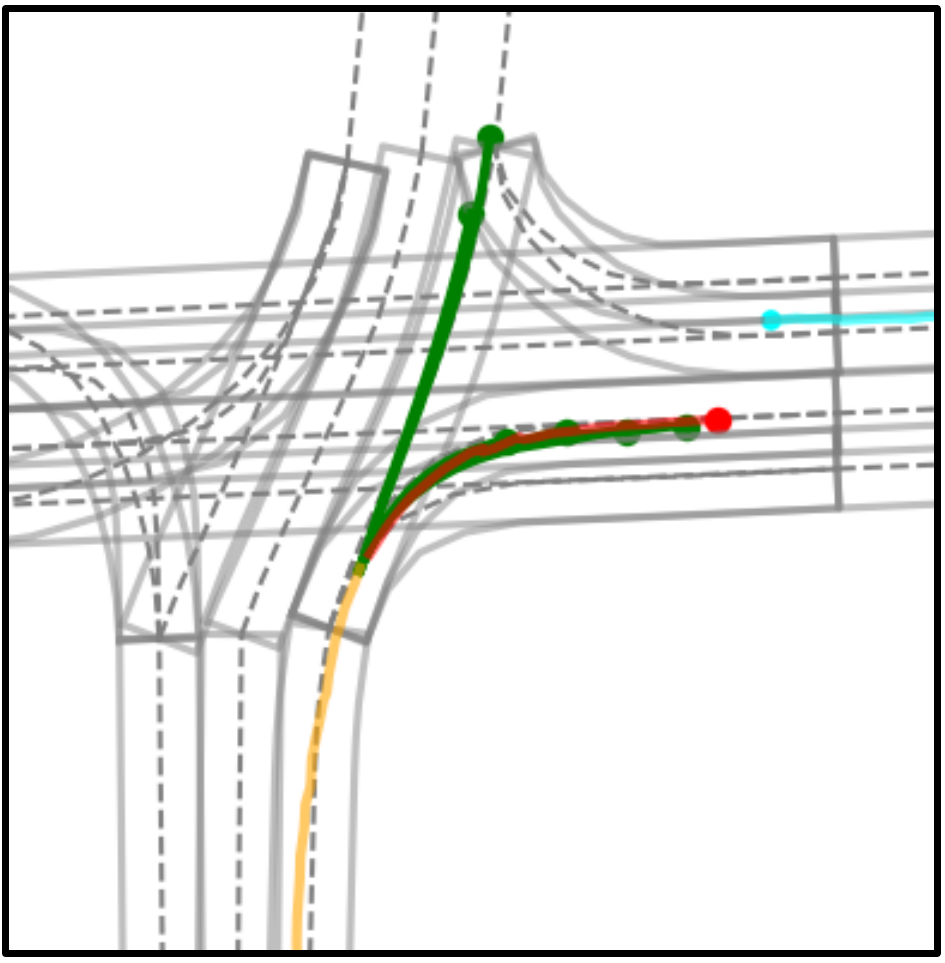}
&\includegraphics[width=0.24\linewidth]{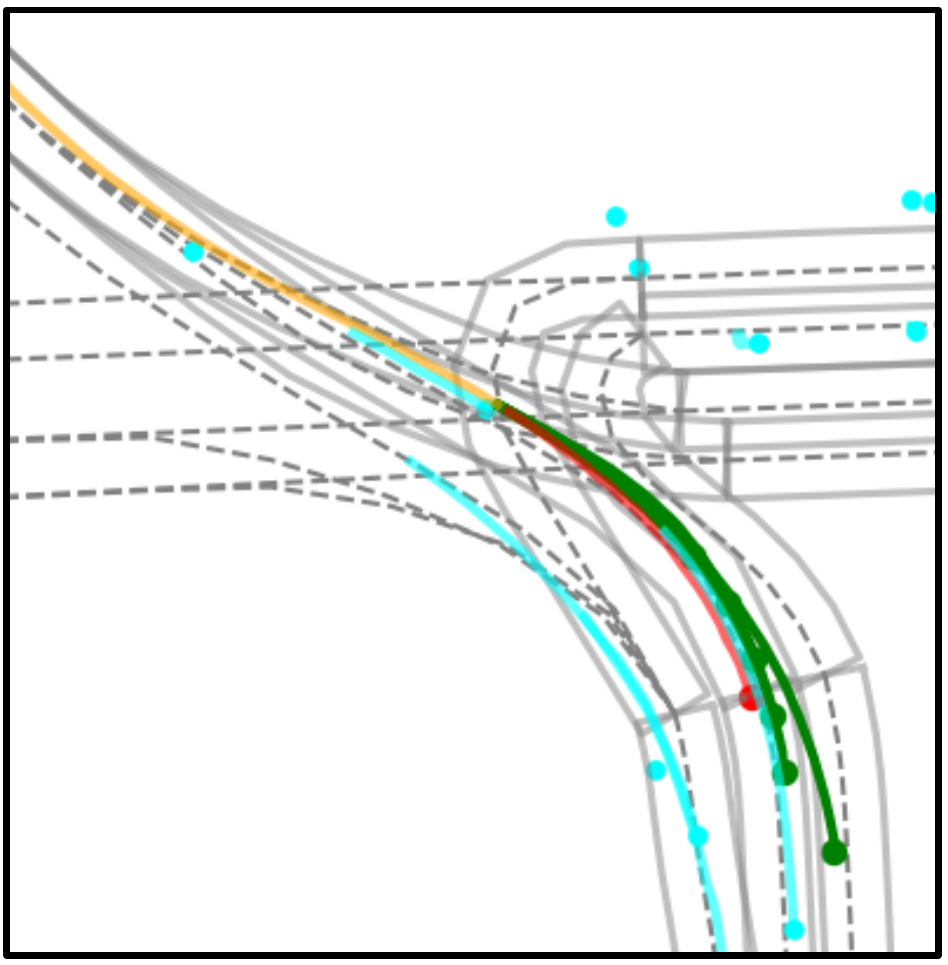}\\
\includegraphics[width=0.24\linewidth]{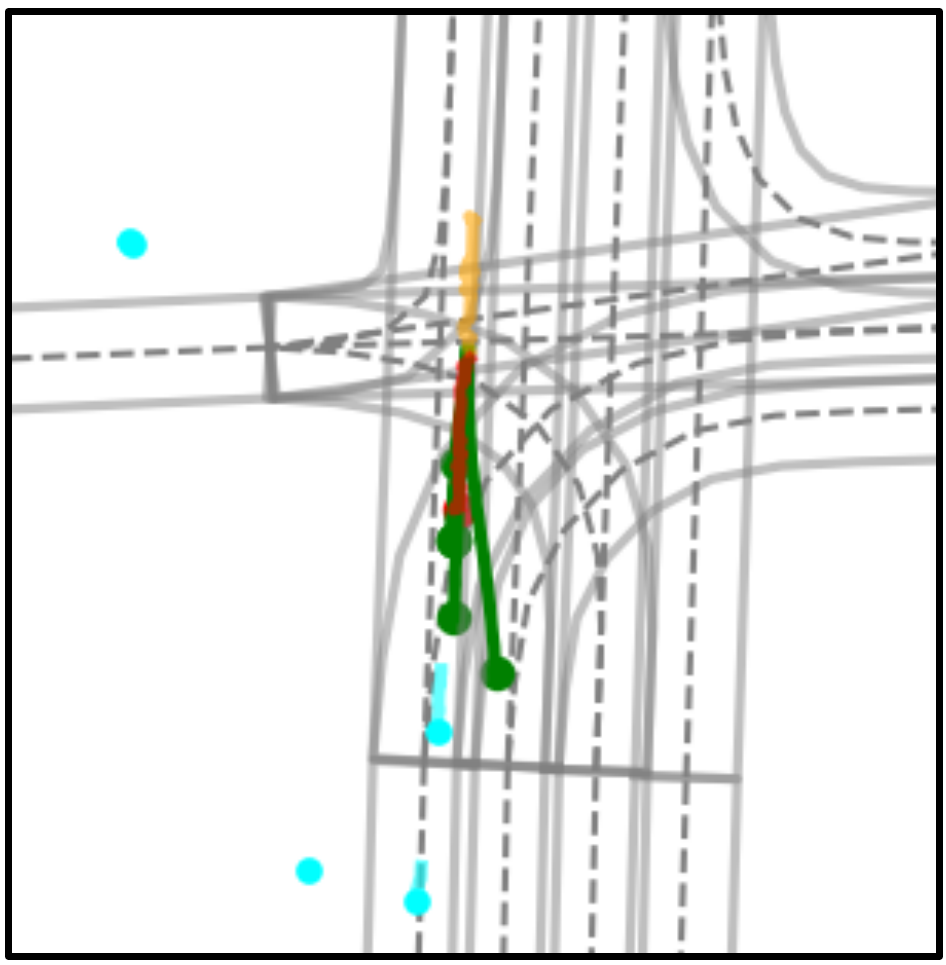}
&\includegraphics[width=0.24\linewidth]{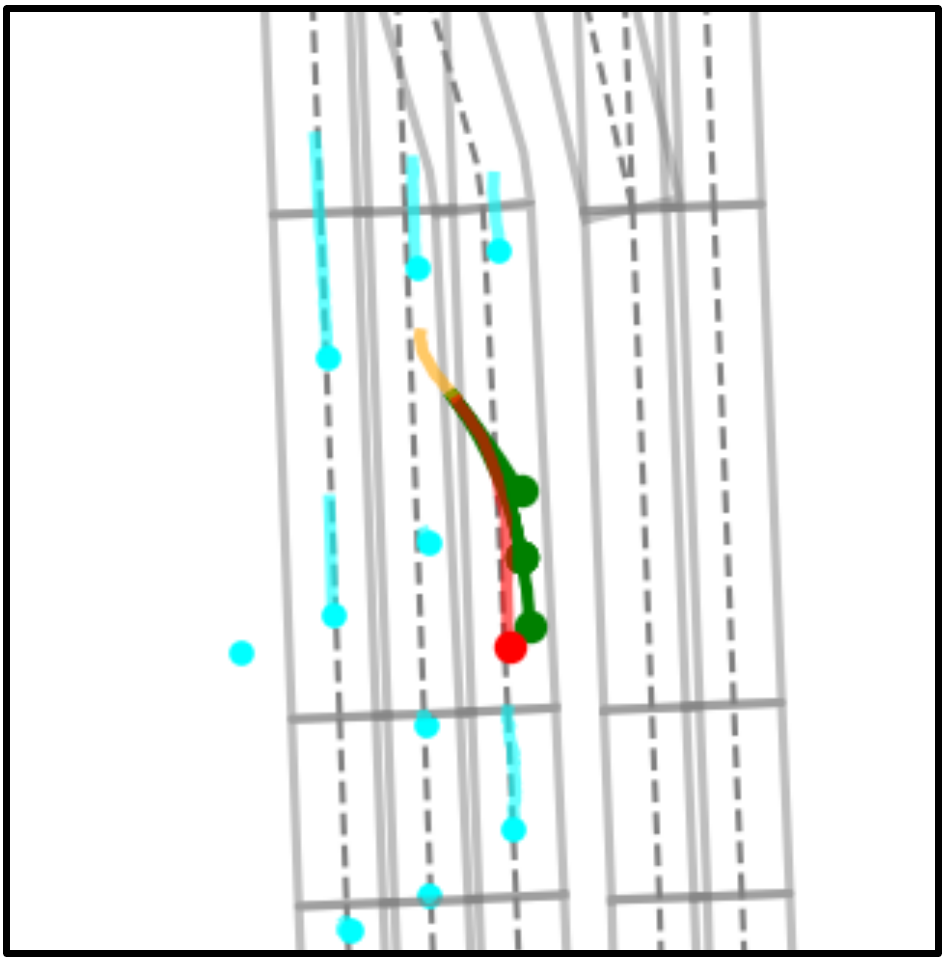}
&\includegraphics[width=0.24\linewidth]{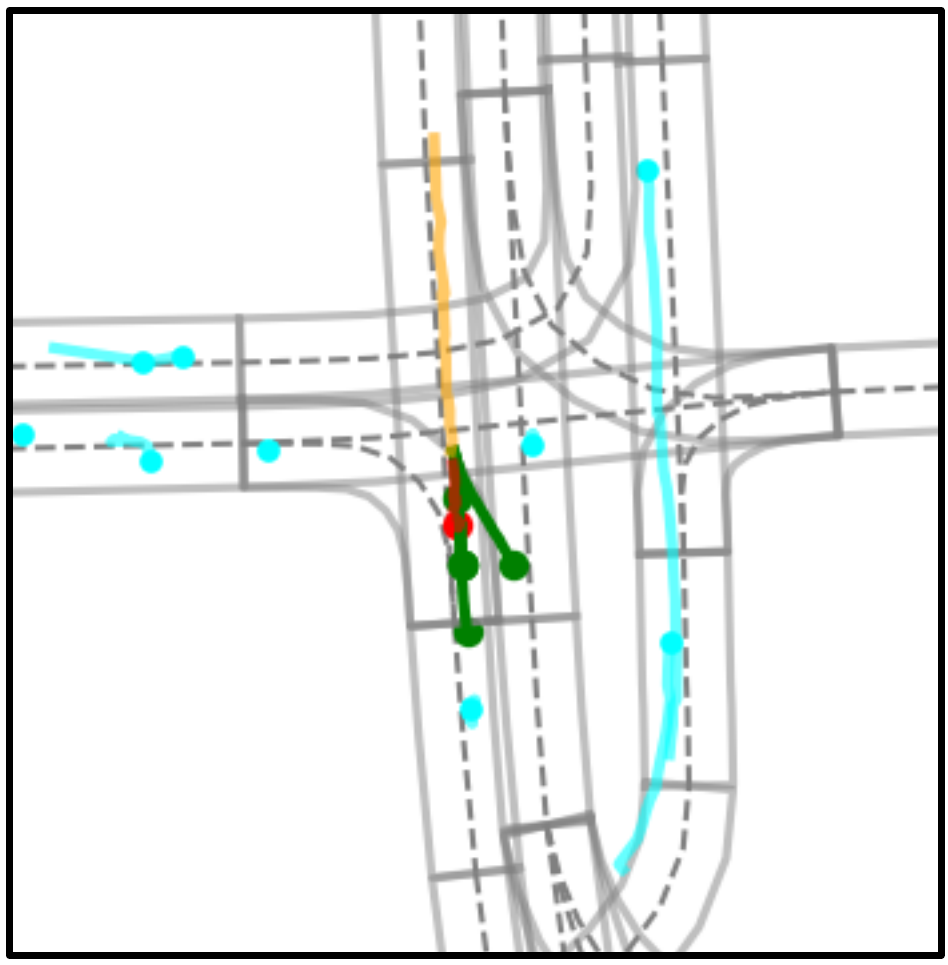}
&\includegraphics[width=0.24\linewidth]{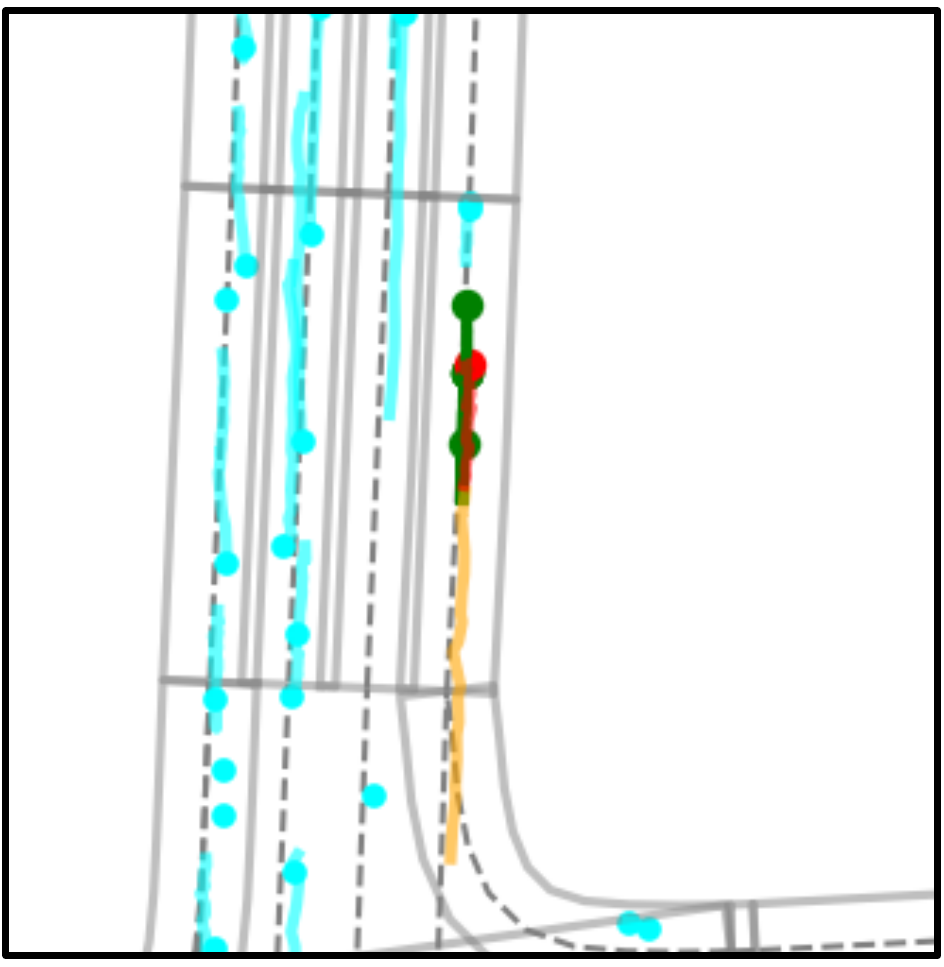}\\
\includegraphics[width=0.24\linewidth]{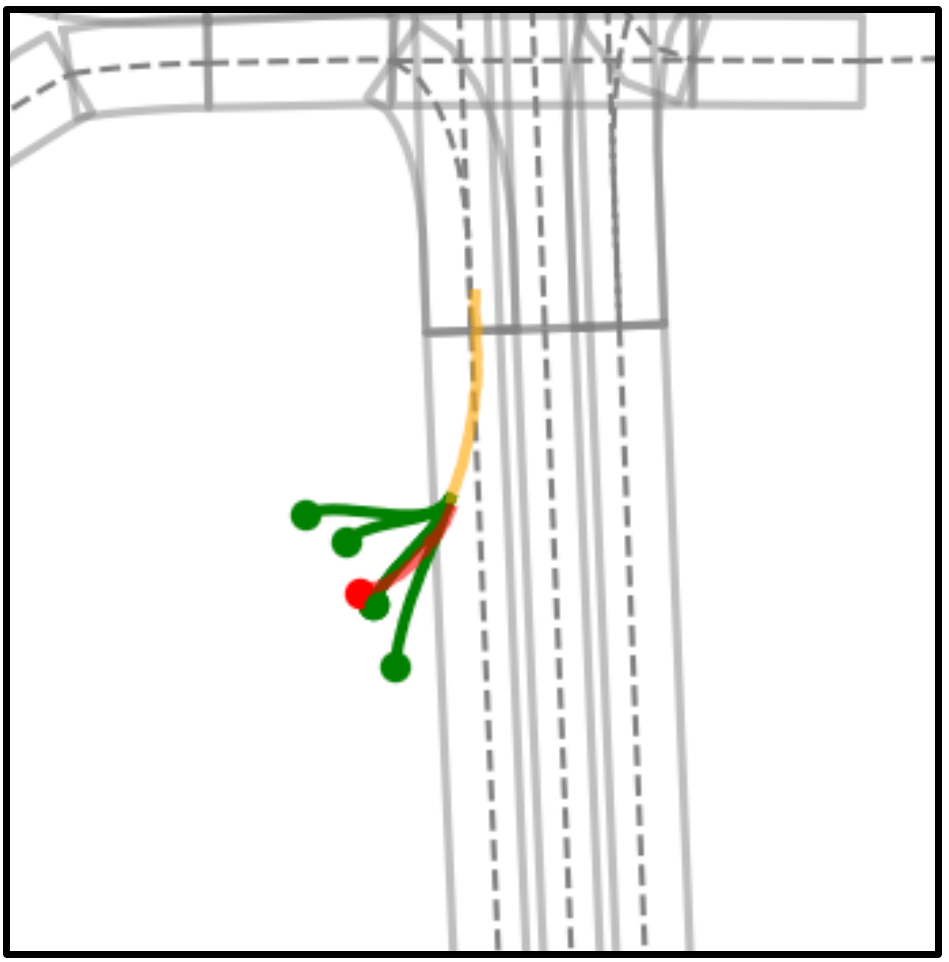}
&\includegraphics[width=0.24\linewidth]{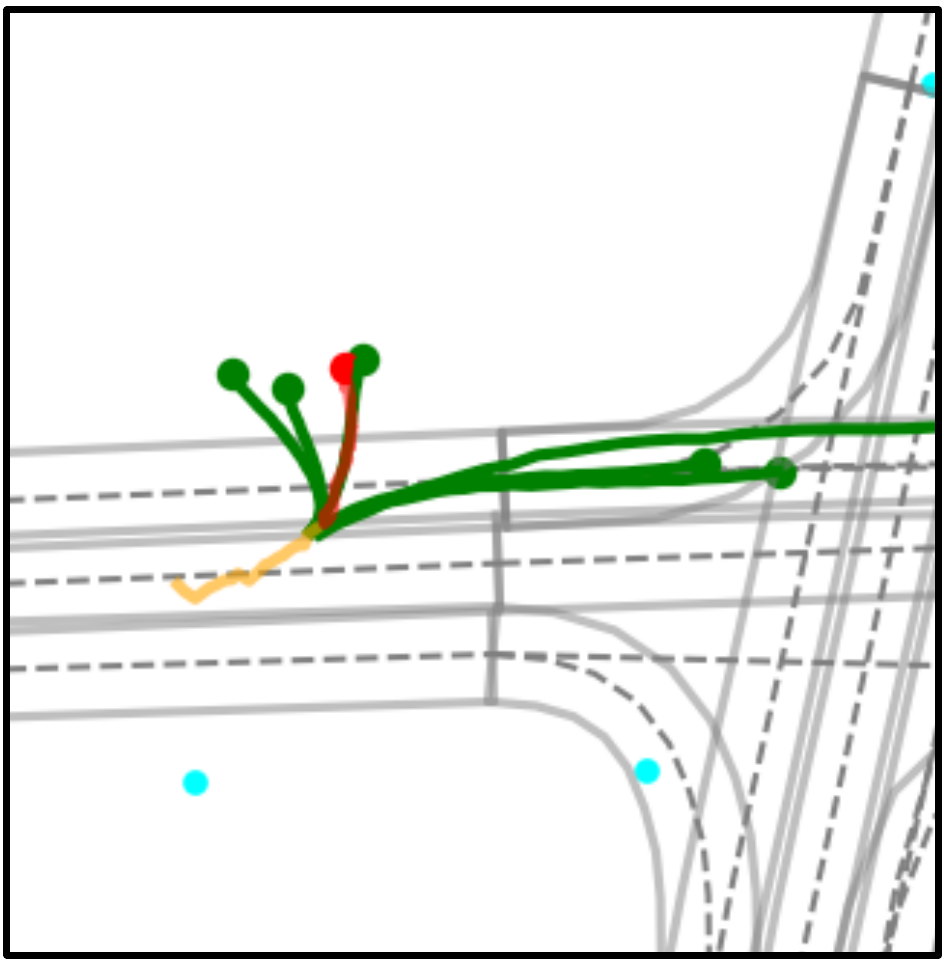}
&\includegraphics[width=0.24\linewidth]{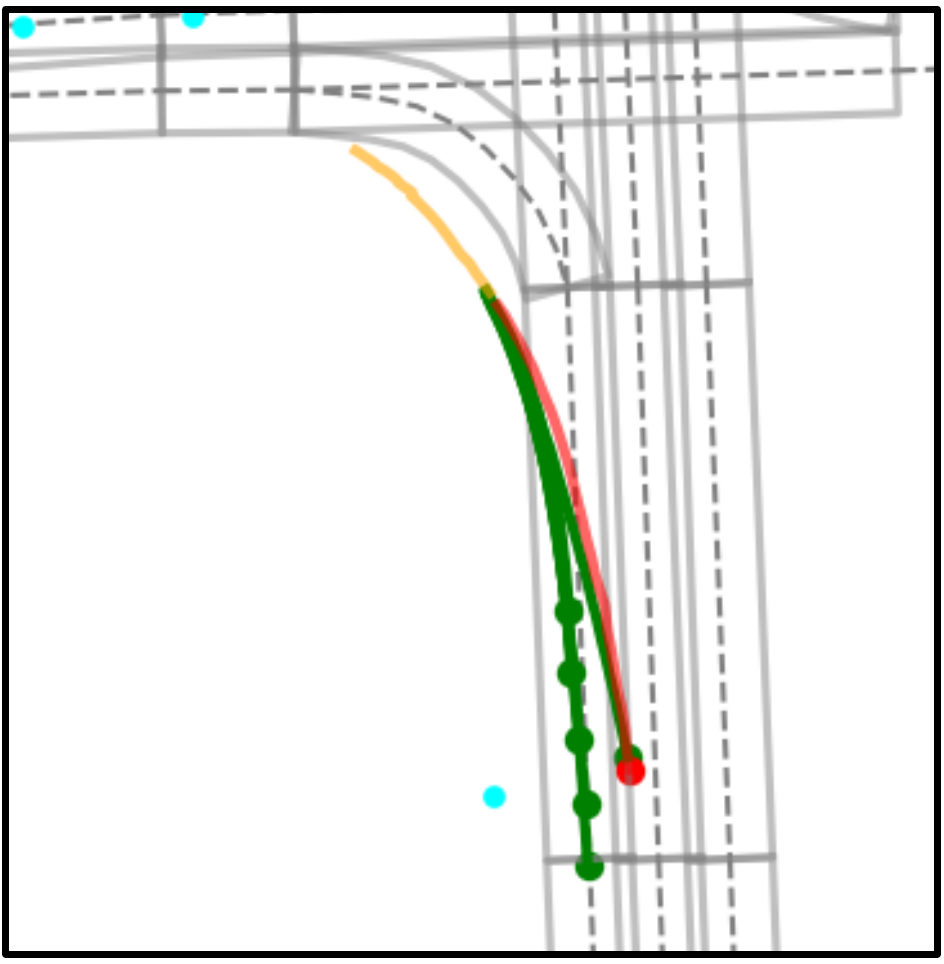}
&\includegraphics[width=0.24\linewidth]{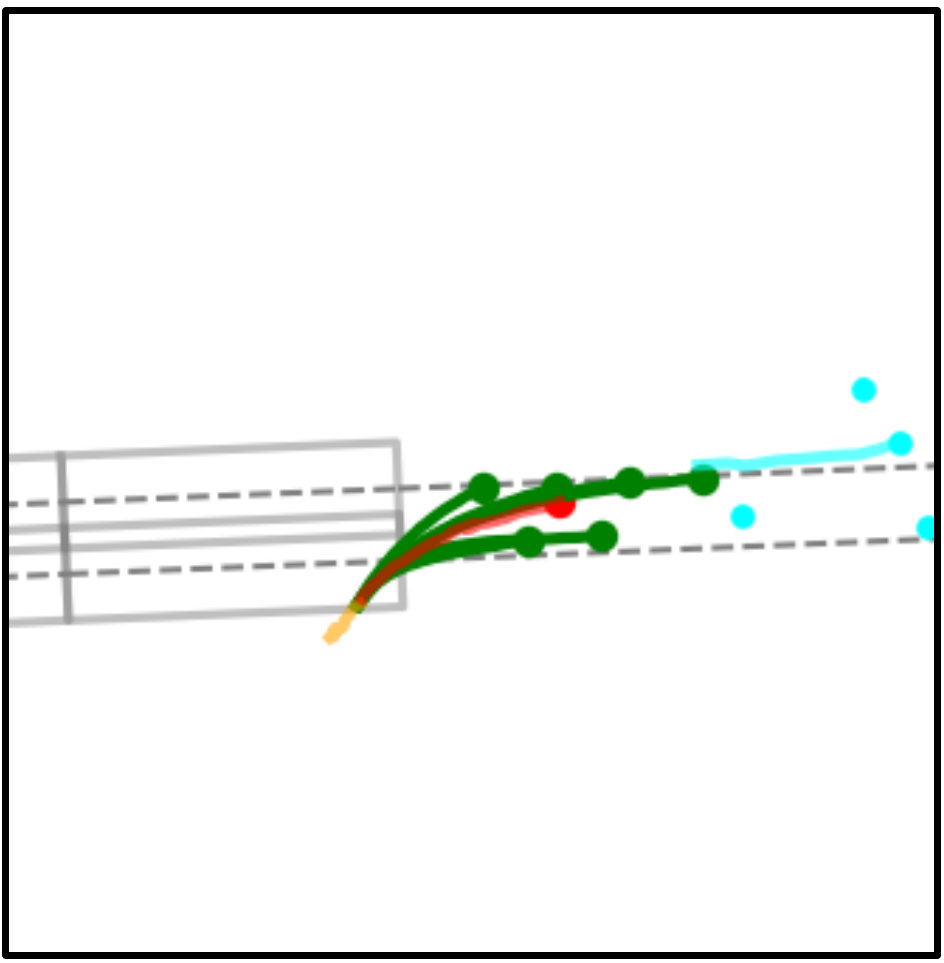}\\

\end{tabular}
\caption{Qualitative results on Argoverse validation set. We show a various of
scenarios including, turning (row 1-2), curved roads (row 3), breaking and
overtaking (row 4), abnormal behaviors (row 5).}
\label{fig:supp_vis}
\end{figure*}

\section{Qualitative Results}
\label{sec:supp_qual}
Lastly, we provide more visualization of our model outputs, as we believe the
metric numbers can only reveal part of the story. We present various
scenarios in Fig.~\ref{fig:supp_vis}, including turning, following curved roads,
interacting with other actors and some abnormal behaviors. On the first two
rows, we show turning behaviors under various map topologies. We can see
our motion forecasting results nicely capture possible modes: turning into
different directions or occupying different lanes after turning. We also do well 
even if the actor is not strictly following the centerlines. On the
third row, we show predictions following curved roads, which are difficult
scenarios for auto-regressive predictors \cite{nmp,intentnet}. On the fourth
row, we show that our model can predict breaking or overtaking behaviors when
leading vehicles are blocking the ego-actor. Finally, we show in the fifth row
that our model also works well when abnormal situations occur, \eg, driving out of
roads. This is impressive as the model relies on
lanes to make predictions, yet it shows capabilities to predict non-map-compliant
behaviors.

\end{document}